\documentclass[lettersize,journal]{IEEEtran}

\usepackage{hyperref}       
\usepackage{url}            

\usepackage[utf8]{inputenc} 
\usepackage{graphicx}
\usepackage{subcaption} 

\usepackage[table]{xcolor} 
\usepackage{xcolor}
\usepackage{amsmath}
\usepackage{amssymb}
\usepackage{multirow}
\usepackage{pifont}
\usepackage{pgffor}      
\usepackage{makecell}
\usepackage{enumitem}
\usepackage{xcolor}

\usepackage{booktabs}       
\usepackage{amsfonts}    
\usepackage{nicefrac}    
\usepackage{microtype}      
\usepackage{adjustbox} 

\def\etal{\emph{et al}}

\definecolor{BlueGreen}{HTML}{008080}
\definecolor{OliveGreen}{RGB}{128, 128, 0}

\newcommand{\reddown}[1]{${\color{red}\downarrow #1}$}

\definecolor{mydarkdarkgreen}{RGB}{93, 150, 74} 
\definecolor{mydarkgreen}{RGB}{216, 233, 199}  
\definecolor{mylightgreen}{RGB}{245, 249, 241}

\definecolor{RedOrange}{RGB}{255, 165, 0}

\definecolor{lightgray}{gray}{.9}
\newcolumntype{I}{!{\vrule width 1pt}}
\makeatletter
\newcommand{\thickhline}{%
    \noalign {\ifnum 0=`}\fi \hrule height 1pt
    \futurelet \reserved@a \@xhline
}
\makeatother

\begin{document}

\title{Stereo Anything: Unifying Zero-shot Stereo Matching with Large-Scale Mixed Data}

\author{\normalsize{
Xianda Guo$*$, Chenming Zhang$*$, Youmin Zhang, Ruilin Wang, Dujun Nie, Wenzhao Zheng\\ Matteo Poggi, Hao Zhao, Mang Ye, Qin Zou, Long Chen \\
\url{https://github.com/XiandaGuo/OpenStereo} \\ 
\IEEEcompsocitemizethanks{
\IEEEcompsocthanksitem Xianda Guo, Mang Ye, and Qin Zou are with the School of Computer Science, Wuhan University, Wuhan, China.
E-mail: \href{mailto:xianda_guo@163.com}{xianda\_guo@163.com}
\IEEEcompsocthanksitem Chenming Zhang is with the Institute of Artificial Intelligence and Robotics, Xi’an Jiaotong University, Xi'an, China.
\IEEEcompsocthanksitem Ruilin Wang and Dujun Nie are with the Institute of Automation, Chinese Academy of Sciences, Beijing, China.
\IEEEcompsocthanksitem Youmin Zhang and Matteo Poggi are with the University of Bologna, Bologna, Italy.
\IEEEcompsocthanksitem Wenzhao Zheng, and Hao Zhao are with Tsinghua University, Beijing, China. 
\IEEEcompsocthanksitem Long Chen is with the Institute of Automation, Chinese Academy of Sciences, Beijing, China, also with the Institute of Artificial Intelligence and Robotics, Xi’an Jiaotong University, Xi'an, China, and also with Waytous Inc., Qingdao, China.
\IEEEcompsocthanksitem $*$: These authors contributed equally to this work. Corresponding authors: Qin Zou and Long Chen. 
}}}

\maketitle

\begin{abstract}
Stereo matching serves as a cornerstone in 3D vision, aiming to establish pixel-wise correspondences between stereo image pairs for depth recovery. Despite remarkable progress driven by deep neural architectures, current models often exhibit severe performance degradation when deployed in unseen domains, primarily due to the limited diversity of training data. In this work, we introduce \textbf{StereoAnything}, a data-centric framework that substantially enhances the zero-shot generalization capability of existing stereo models. Rather than devising yet another specialized architecture, we scale stereo training to an unprecedented level by systematically unifying heterogeneous stereo sources: (1) curated labeled datasets covering diverse environments, and (2) large-scale synthetic stereo pairs generated from unlabeled monocular images. Our mixed-data strategy delivers consistent and robust learning signals across domains, effectively mitigating dataset bias. Extensive zero-shot evaluations on four public benchmarks demonstrate that Stereo Anything achieves state-of-the-art generalization. This work paves the way towards truly universal stereo matching, offering a scalable data paradigm applicable to any stereo image pair.
We extensively evaluate the zero-shot capabilities of our model on four public datasets,  showcasing its impressive ability to generalize to any stereo image pair. 

\end{abstract}

\section{Introduction}
\label{sec:intro}

Computer vision is currently undergoing a revolution due to the development of foundation models in object recognition~\cite{wang2023detecting}, image segmentation~\cite{kirillov2023segment}, and depth estimation~\cite{guo2023simple, depthanytingv1, Marigold, depthanytingv2}, which demonstrate strong zero- and few-shot performance across various downstream tasks. Stereo matching is fundamental for enabling depth perception and 3D reconstruction of observed scenes, playing a critical role in applications such as robotics~\cite{linestereo2015}, autonomous driving~\cite{guo2025lightstereo, guo2023openstereo, jing2024matchstereovideos, jing2024matchsv}, and augmented reality~\cite{smart_iotj2020}. However, the exploration of foundation models in stereo matching remains limited due to the extreme difficulty of obtaining accurate disparity ground truth (GT) data. Although numerous stereo datasets~\cite{sceneflow,kitti2012,kitti2015,yang2019drivingstereo,karaev2023dynamicstereo} have been published, their scale, diversity, and annotation quality remain insufficient to support the development of an ideal stereo foundation model. Even when combined, existing datasets do not fully capture the breadth of real-world scenarios needed for robust generalization. Furthermore, efficiently integrating and utilizing these datasets for training remains a challenge due to disparities in domain characteristics, data formats, and distribution shifts.

\begin{figure}
    \centering
    \includegraphics[width=1\linewidth]{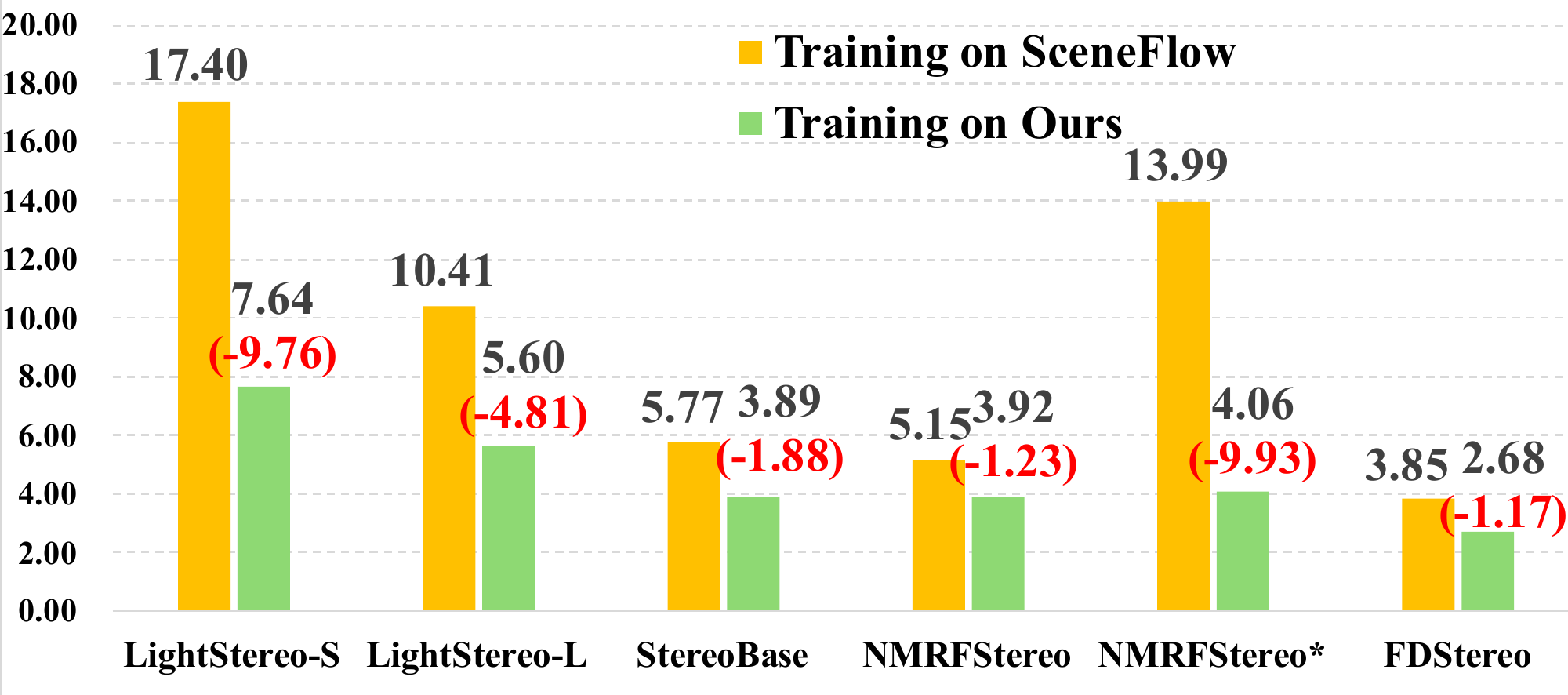}
    \caption{Comparison of average errors when training on SceneFlow versus our mixed-data corpus.}
    \label{fig:teaser}
\end{figure}

A few recent efforts~\cite{croco_v2, cfnet2021} have attempted to address this challenge. 
CroCo-Stereo \cite{croco_v2} is trained on a combination of five datasets, while CFNet \cite{cfnet2021} is fine-tuned on the four datasets. 
However, the selection rationale behind these datasets is not rigorously analyzed, leaving open questions about their contribution to model generalization and performance consistency.
Another promising direction is introduced with Stereo-from-mono~\cite{stereo-from-mono}, which attempts to alleviate data scarcity by synthesizing stereo image pairs and disparity maps from monocular images. However, this approach resulted in the creation of only 500,000 data samples, which is relatively limited, considering the scale required to train foundation models effectively. While this effort represents an important step towards reducing the dependency on expensive stereo data collection, it is still insufficient for building large-scale models capable of generalizing well to diverse real-world conditions.
These observations raise two fundamental questions: \textbf{\textit{(1) How can we fully exploit the existing labeled stereo data? (2) How can monocular foundation models be leveraged to generate pseudo-stereo data at scale to supplement stereo training?}}

In this paper, we introduce Stereo Anything, a scalable and generalizable stereo matching framework that bridges these two directions. Rather than focusing on architectural innovations, we propose a data-centric approach that unifies diverse data sources into a large-scale training corpus. Specifically, we utilize numerous publicly available annotated stereo datasets as a core part of our training data to further increase the quality, quantity, and variety of training data available.
Furthermore, inspired by the successes of the stereo-from-mono~\cite{stereo-from-mono} approach and monocular depth foundation models~\cite{midas, depthanytingv1, Marigold, depthanytingv2}, we supplement our training with a large volume of synthetic stereo data generated from monocular images. As shown in Fig.~\ref{fig:teaser}, StereoAnything empowers any existing stereo matching models--from lightweight architectures~\cite{guo2025lightstereo} to accuracy-focused designs~\cite{NMRFStereo,xu2023iterative}--to achieve strong zero-shot performance across four public benchmarks.

Our contributions can be summarized as follows:
\begin{itemize} 
\item We propose StereoAnything, a data-centric framework that unifies diverse labeled stereo datasets and pseudo-stereo data synthesized from monocular images to enhance zero-shot generalization for stereo matching.
\item We emphasize the importance of scaling up labeled stereo datasets through an exhaustive study of how different stereo datasets affect the performance of trained stereo models.
\item We propose a novel pipeline for synthesizing stereo images from monocular images. We further scale up the training data by effectively incorporating both labeled stereo data and diverse unlabeled monocular images for training stereo networks. 
\item Our final training strategy empowers any stereo model to significantly enhance its zero-shot generalization capabilities.
\end{itemize}

\section{Related Work}

Stereo matching has been extensively studied in the last decade.
As shown in Fig.~\ref{fig:pipeline}, 
Stereo matching based on deep learning involves estimating disparity from a pair of rectified stereo images. It mainly consists of four essential components~\cite{psmnet2018, NMRFStereo, tosi2024survey}: feature extraction, cost construction, cost aggregation, and disparity regression.
We identify and review two main bodies of literature, concerning in-domain and cross-domain stereo matching, as well as review techniques for multi-dataset training.

\textbf{In-domain Stereo Matching.}
Early methods depended on hand-crafted features and optimization-based matching~\cite{scharstein2002taxonomy, hirschmuller2007sgm}. With the rise of deep learning, the CNN-based model~\cite{gcnet, psmnet2018, gwcnet2019, dsmnet2020} has revolutionized stereo matching by learning feature representations and matching costs directly from data, leading to significant improvements in accuracy and robustness.
GCNet~\cite{gcnet} firstly constructs concatenation-based 4D cost volumes and utilizes 3D CNNs for cost aggregation.
Based on GCNet, PSMNet~\cite{psmnet2018} enhances feature extraction using a spatial pyramid pooling module and employs a stacked hourglass architecture for 3D cost aggregation.
GANet~\cite{ganet2019} accelerates the network by replacing the 3D CNN with a semi-global aggregation layer and a locally guided aggregation layer.
LEAStereo~\cite{leastereo} employs a neural architecture search to optimize stereo-matching networks. 
There has been a growing interest in video stereo matching~\cite{zhang2022temporalstereo,karaev2023dynamicstereo,zeng2024temporally,jing2024matchstereovideos, jing2024matchsv}, particularly in leveraging temporal information to achieve consistent stereo matching.
Recently, DEFOM-Stereo~\cite{defomstereo} and Monster~\cite{cheng2025monster} integrate monocular depth estimation models into stereo matching frameworks to enhance disparity prediction.
Despite their success, these approaches often exhibit domain-specific biases, limiting their performance on unseen datasets.

\textbf{Cross-domain Stereo Matching.}
Domain-invariant representations are crucial for cross-domain stereo matching.
Tonioni~\etal~\cite{tonioni2017unsupervised} propose an unsupervised adaptation method that can fine-tune a stereo-matching model without any ground-truth information. 
DSMNet~\cite{dsmnet2020} incorporates a trainable non-local graph filter and domain normalization layer to extract robust structural features. CFNet~\cite{cfnet2021} addresses the large domain difference by introducing a cascade and fused cost volume.
FCStereo~\cite{zhang2022revisiting} proposes contrastive learning to maintain feature consistency between matching pixels, which is vital for the generalization of stereo matching.
To reduce domain shifts, Chang~\etal~\cite{chang2023domain} presents a hierarchical visual transformation network that focuses on learning shortcut-invariant representation from synthetic data.
RAFT-Stereo~\cite{raftstereo} and IGEV~\cite{xu2023iterative} 
introduce multi-level ConvGRUs to update the disparity map and achieve strong cross-dataset generalization iteratively. 
MS-Net~\cite{cai2020matching}, GraftNet~\cite{liu2022graftnet}, and FormerStereo~\cite{zhang2025learning} obtain domain-invariant features by leveraging external priors and employing robust feature extractors. 
NMRF-Stereo~\cite{NMRFStereo} presents a neural MRF model, which exhibits strong cross-domain generalization and can recover sharp edges.
FoundationStereo~\cite{wen2025foundationstereo} and Stereo Anywhere \cite{Bartolomei_2025_CVPR} propose a zero-shot generalizable stereo matching model by adapting a ViT-based monocular depth estimation model~\cite{depthanytingv2} into the stereo matching. Despite their remarkable cross-dataset generalization capability, this approach suffers from substantial computational overhead, making it highly time-consuming in practical scenarios.

We refer the reader to recent surveys \cite{laga2020survey,poggi2021synergies,tosi2024survey} for further details on both in-domain and cross-domain stereo.

\begin{table*}[t]
\caption{\textbf{Datasets used in our work.} Top: Our training sets. Bottom: Our test sets. In/out/Dy/W/Acc./Divers. refers to Indoor/Outdoor/Dynamic/Weather/Accuracy/Diversity. FL refers to focal length. Range refers to disparity range. Ave./Med. refers to the average/median of disparity.}
\label{tab:datasets_summary}
\centering
\scriptsize
\setlength\tabcolsep{6pt}
\renewcommand\arraystretch{1.1}
\resizebox{\linewidth}{!}{
\begin{tabular}{l|c|cccccc|cc|c|c|c|c|c|c}
\thickhline 
\rowcolor{mydarkgreen} 
\textbf{Dataset} & \textbf{Year} & In & Out & Dy & Video & Dense & W & Acc. & Divers. & Type & Resolution & Baseline & FL & Range & Ave./Med. \\
\hline\hline
Sintel~\cite{Sintel} & ECCV12 & \checkmark & \checkmark & \checkmark & \checkmark & \checkmark & \ding{55} & High & Med & Syn & 1024×436 & 0.1m & - & 0–972 & 66.5/25 \\
\rowcolor{mylightgreen} 
SceneFlow~\cite{sceneflow} & CVPR16 & \checkmark & \checkmark & \checkmark & \checkmark & \checkmark & \ding{55} & High & High & Syn & 960×540 & 0.54m & - & 0–10501 & 53.9/36 \\
FallingThings~\cite{tremblay2018falling} & CVPRW18 & \checkmark & \checkmark & \ding{55} & \ding{55} & \checkmark & \ding{55} & High & Low & Syn & 960×540 & 6cm & 768.2px & 7–461 & 35.2/34 \\
\rowcolor{mylightgreen} 
VKITTI2~\cite{cabon2020virtual} & ArXiv20 & \ding{55} & \checkmark & \checkmark & \checkmark & \checkmark & \ding{55} & High & Mid & Syn & 1242×375 & 0.54m & 725px & 0–411 & 30.1/25 \\
\rowcolor{mylightgreen} 
TartanAir~\cite{wang2020tartanair} & IROS20 & \checkmark & \checkmark & \checkmark & \checkmark & \checkmark & \ding{55} & High & High & Syn & 640×480 & – & – & 0–499 & 21.0/13 \\
InStereo2K~\cite{bao2020instereo2k} & SCIS20 & \checkmark & \ding{55} & \ding{55} & \ding{55} & \ding{55} & \ding{55} & Low & Low & LiDAR & 1080×860 & 10cm & 8mm & 0–328 & 78.4/74 \\
\rowcolor{mylightgreen} 
UnrealStereo4K~\cite{tosi2021smd} & CVPR21 & \checkmark & \checkmark & \ding{55} & \ding{55} & \checkmark & \ding{55} & High & High & Syn & 3840×2160 & 0.2m/0.5m & – & 0–1515 & 175.3/135 \\
CREStereo~\cite{Crestereo} & CVPR22 & \checkmark & \checkmark & \ding{55} & \ding{55} & \checkmark & \ding{55} & High & High & Syn & 1920×1080 & – & – & 0–2048 & 15.2/8 \\
\rowcolor{mylightgreen} 
Spring~\cite{mehl2023spring} & CVPR23 & \ding{55} & \checkmark & \checkmark & \checkmark & \checkmark & \ding{55} & High & Low & Syn & 1920×1080 & 6.5cm & – & 0–554 & 38.1/19 \\
DynamicReplica~\cite{karaev2023dynamicstereo} & CVPR23 & \checkmark & \ding{55} & \checkmark & \checkmark & \checkmark & \ding{55} & High & Low & Syn & 1280×720 & 4–30cm & – & 3–656 & 62.7/48 \\
\rowcolor{mylightgreen} 
FoundationStereo~\cite{wen2025foundationstereo} & CVPR25 & \ding{55} & \checkmark &  \ding{55} & \ding{55}  & \checkmark & \checkmark & High & High & Syn & 1280×720 &- &- &1-700& 52.5/39\\

StereoCarla & ArXiv25 & \ding{55} & \checkmark & \checkmark & \checkmark & \checkmark & \checkmark & High & High & Syn & 1600×900 & 10–300cm & 1385.6px & 0–3318 & 80.8/31 \\

\hline\hline
KITTI12~\cite{kitti2012} & CVPR12 & \ding{55} & \checkmark & \checkmark & \checkmark & \ding{55} & \ding{55} & Med & Low & LiDAR & 1242×375 & 0.54m & 720px & 4–232 & 40.1/38 \\
\rowcolor{mylightgreen} 
Middlebury~\cite{middlebury} & GCPR14 & \checkmark & \ding{55} & \ding{55} & \ding{55} & \ding{55} & \ding{55} & High & Low & LiDAR & ~6Mpx & 140–400mm & 1100–3600px & 15–323 & 72.5/63 \\
KITTI15~\cite{kitti2015} & CVPR15 & \ding{55} & \checkmark & \checkmark & \checkmark & \ding{55} & \ding{55} & Med & Low & LiDAR & 1242×375 & 0.54m & 520px & 4–230 & 35.2/33 \\
\rowcolor{mylightgreen} 
ETH3D~\cite{eth3d} & CVPR17 & \checkmark & \checkmark & \ding{55} & \checkmark & \ding{55} & \ding{55} & High & Low & LiDAR & ~0.3Mpx & 59.5–60.4mm & 529–712px & 0–62 & 13.7/10 \\
\thickhline
\end{tabular}}
\end{table*}

\textbf{Multi-Dataset Training.}
To tackle the challenge of generalization, some researchers have increasingly explored the use of multi-dataset training strategies. By combining multiple datasets, models can be trained to capture diverse scene characteristics better, making them more robust. Although well-established in fields like object recognition~\cite{wang2023detecting}, segmentation~\cite{kirillov2023segment}, and monocular depth estimation~\cite{midas, depthanytingv1, Marigold, depthanytingv2}, the use of this strategy in stereo matching remains relatively underexplored.
In monocular depth estimation, MiDaS~\cite{midas} pioneered a data-centric approach by mixing heterogeneous datasets spanning indoor, outdoor, synthetic, and real-world scenes, and by introducing a scale-and-shift invariant loss to reconcile inconsistent depth ranges across domains. This strategy enabled the model to learn domain-agnostic geometric priors, resulting in strong zero-shot cross-dataset transfer without requiring any fine-tuning. Building on this idea, Depth Anything v1/v2~\cite{depthanytingv1,depthanytingv2} substantially scaled up both the diversity and volume of training data while leveraging high-capacity Vision Transformer backbones to absorb such large-scale supervision. 
In stereo matching, multi-dataset training has been explored only in a limited scope, often by combining a small number of complementary datasets to balance synthetic pretraining and real-world fine-tuning. While such strategies can improve performance on specific benchmarks, they are typically constrained by the narrow diversity of included datasets, limited disparity distributions, and scene biases. Moreover, the lack of a standardized framework for unifying heterogeneous stereo sources—spanning real-world captures, large-scale synthetic datasets, and pseudo-stereo generated from monocular images—has hindered progress toward universal stereo models. This gap motivates our work, where we systematically scale up stereo training data across domains to enable robust zero-shot generalization.
In this work, we introduce StereoAnything, a stereo-matching model that systematically leverages multi-dataset learning. Inspired by the successes in related domains, we aim to enhance the generalization capabilities of stereo-matching models by training on a diverse collection of labeled datasets combined with a large amount of synthetically generated pseudo-stereo data.

\section{StereoAnything}

We now introduce StereoAnything, designed to train any robust stereo network using large-scale data.

\subsection{Stereo Matching}
\begin{figure}[t]
    \centering
    \includegraphics[width=0.49\textwidth]{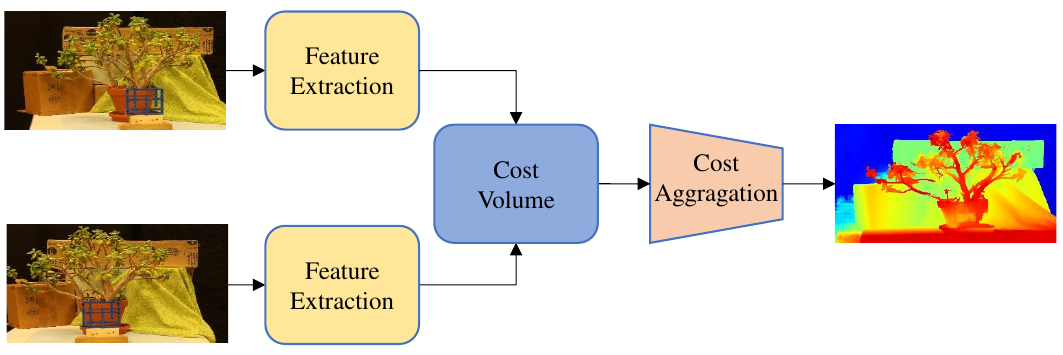}
    \caption{\textbf{Stereo Matching Pipeline}. Deep stereo models usually perform feature extraction, cost construction, cost aggregation, and disparity regression. }  
    \label{fig:pipeline}
\end{figure}

As illustrated in Fig.~\ref{fig:pipeline}, the stereo matching pipeline typically involves four stages: feature extraction, cost construction, cost aggregation, and disparity regression. In the first stage, feature extraction is performed on both the left and right images to capture key visual information that can be used to identify corresponding points between the images. The extracted features are then used to build a cost volume, which represents the similarity between each pixel in the left image and potential matches in the right image at different disparity levels.
Once the cost volume is constructed, cost aggregation is applied to refine the volume by considering the contextual information from neighboring pixels, reducing the impact of noise and improving matching accuracy, especially in occluded or textureless regions. Finally, disparity regression is applied to the aggregated cost volume to generate a disparity map, which provides depth information for each pixel in the image. 
Modern deep learning-based stereo matching models~\cite{NMRFStereo,guo2023openstereo} have enhanced this pipeline by using CNNs or other advanced architectures like the Vision Transformer (ViT) to perform end-to-end learning for both feature extraction and cost volume optimization. These models are trained jointly, optimizing the entire pipeline to learn both local and global matching constraints. 

Overall, stereo matching has evolved from a simple pixel-wise comparison task to a complex, deep learning-driven process that optimizes feature extraction, cost aggregation, and disparity regression jointly. The ongoing research continues to push the boundaries of accuracy and generalization, making stereo matching an essential technology for applications in autonomous driving, robotics, and 3D scene reconstruction.

\begin{figure*}[t!]
    \centering
    \begin{subfigure}[b]{0.16\textwidth}
        \centering
        \includegraphics[width=\textwidth]{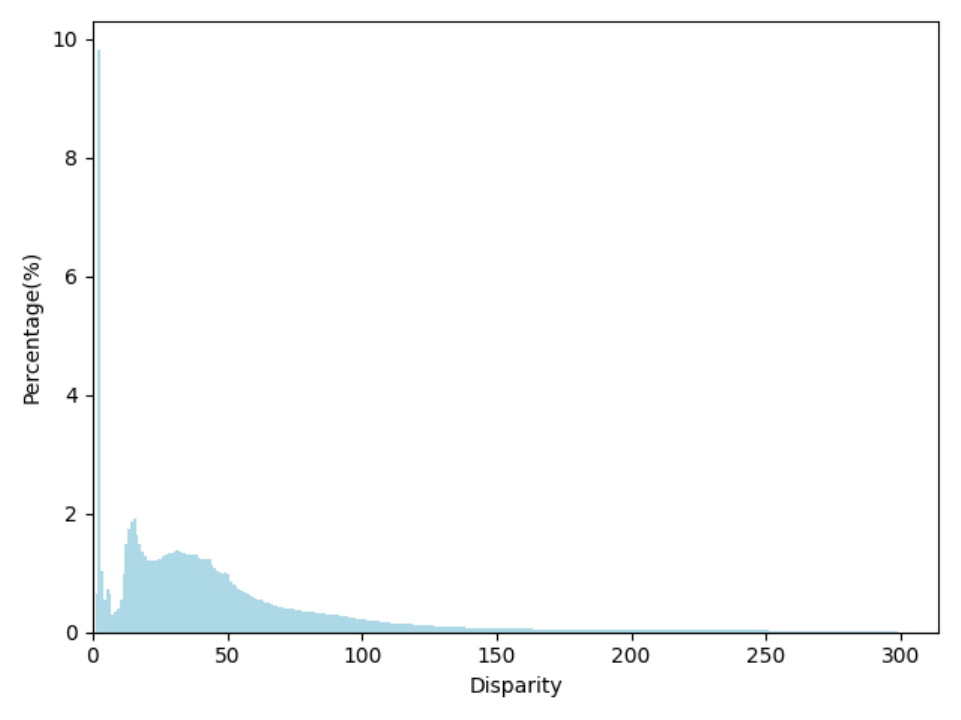}
        \caption{SceneFlow}
    \end{subfigure}
    \begin{subfigure}[b]{0.16\textwidth}
        \centering
        \includegraphics[width=\textwidth]{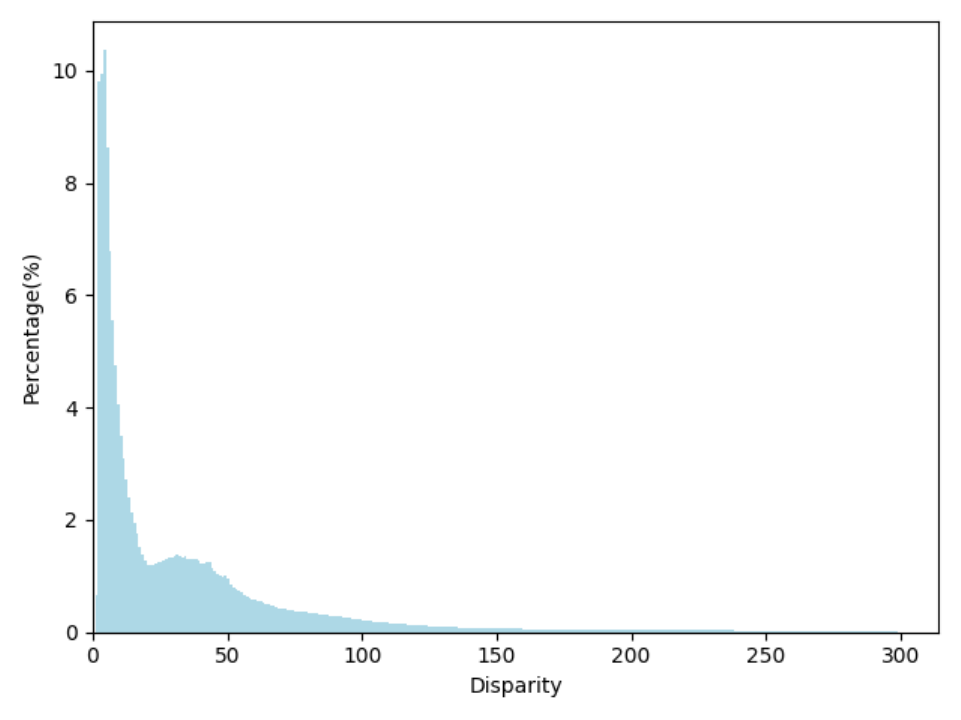}
        \caption{CREStereo}
    \end{subfigure}
    \begin{subfigure}[b]{0.16\textwidth}
        \centering
        \includegraphics[width=\textwidth]{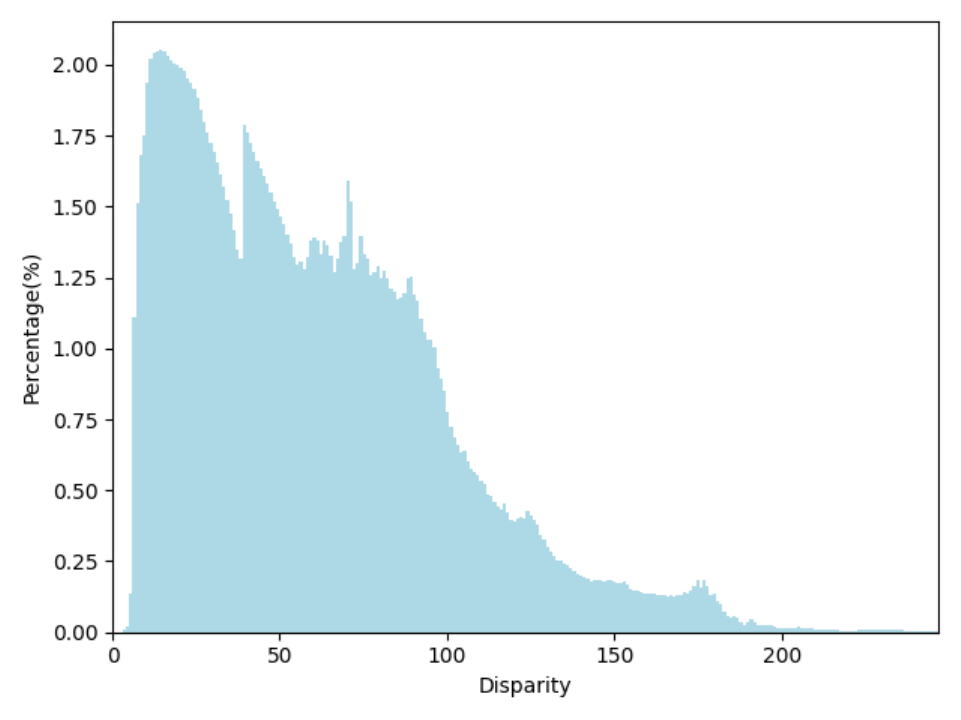}
        \caption{InStereo2K}
    \end{subfigure}
    \begin{subfigure}[b]{0.16\textwidth}
        \centering
        \includegraphics[width=\textwidth]{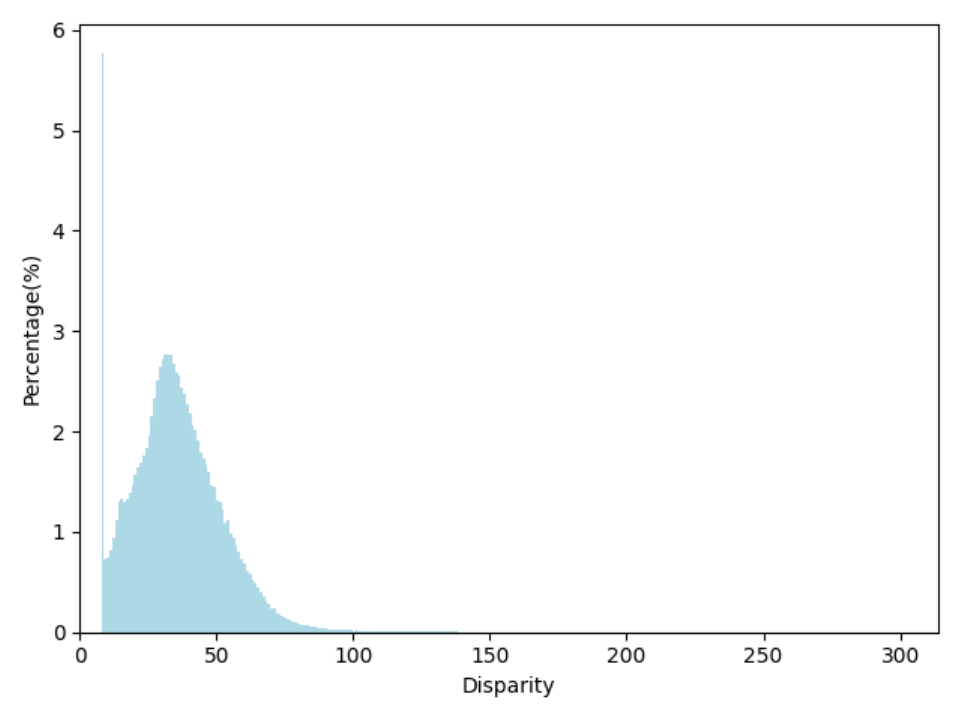}
        \caption{FallingThings}
    \end{subfigure}
    \begin{subfigure}[b]{0.16\textwidth}
        \centering
        \includegraphics[width=\textwidth]{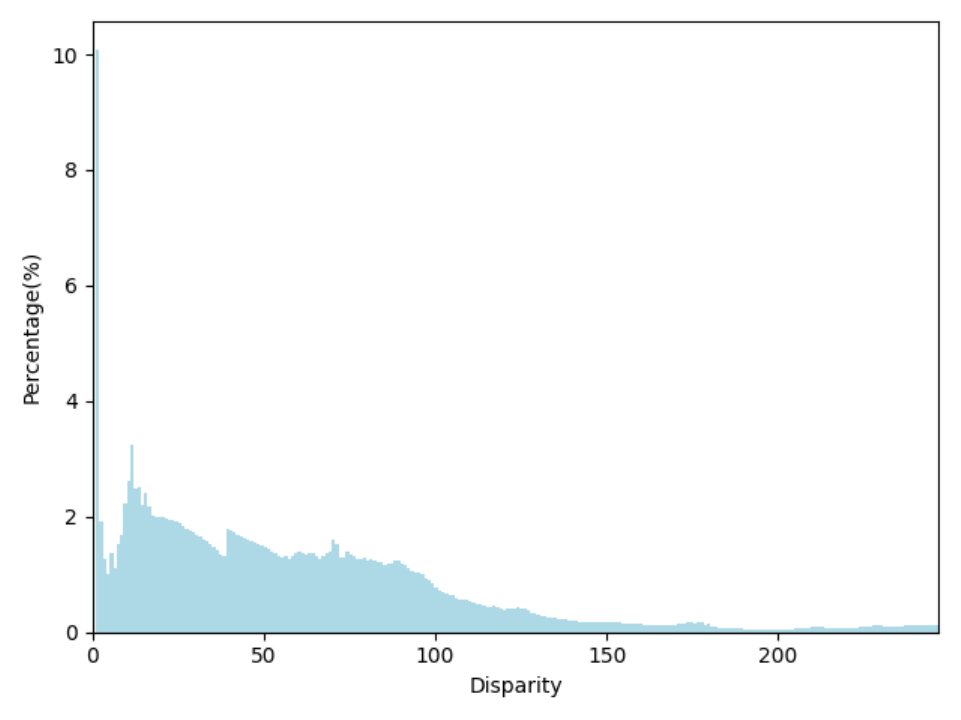}
        \caption{Sintel}
    \end{subfigure}
    \begin{subfigure}[b]{0.16\textwidth}
        \centering
        \includegraphics[width=\textwidth]{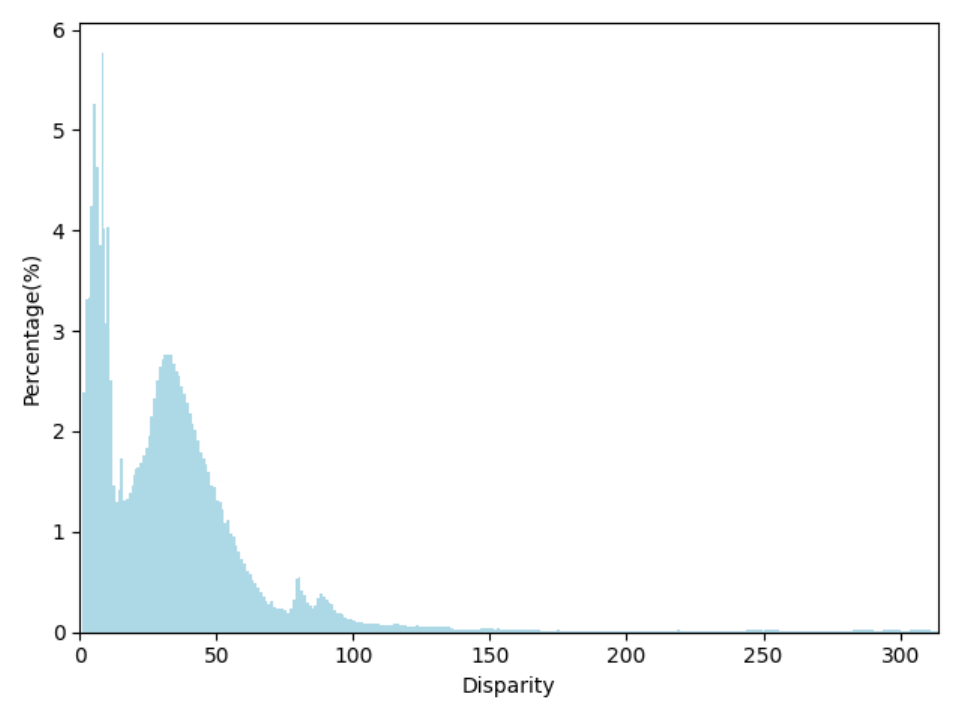}
        \caption{Spring}
    \end{subfigure}
    
    \begin{subfigure}[b]{0.16\textwidth}
        \centering
        \includegraphics[width=\textwidth]{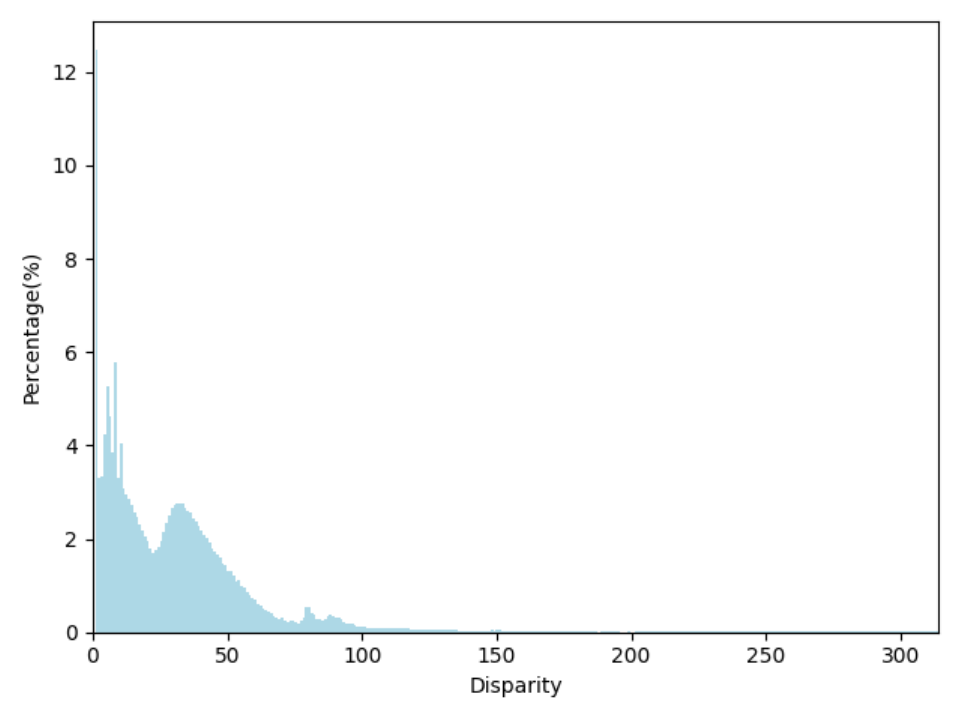}
        \caption{Tartanair}
    \end{subfigure}
    \begin{subfigure}[b]{0.16\textwidth}
        \centering
        \includegraphics[width=\textwidth]{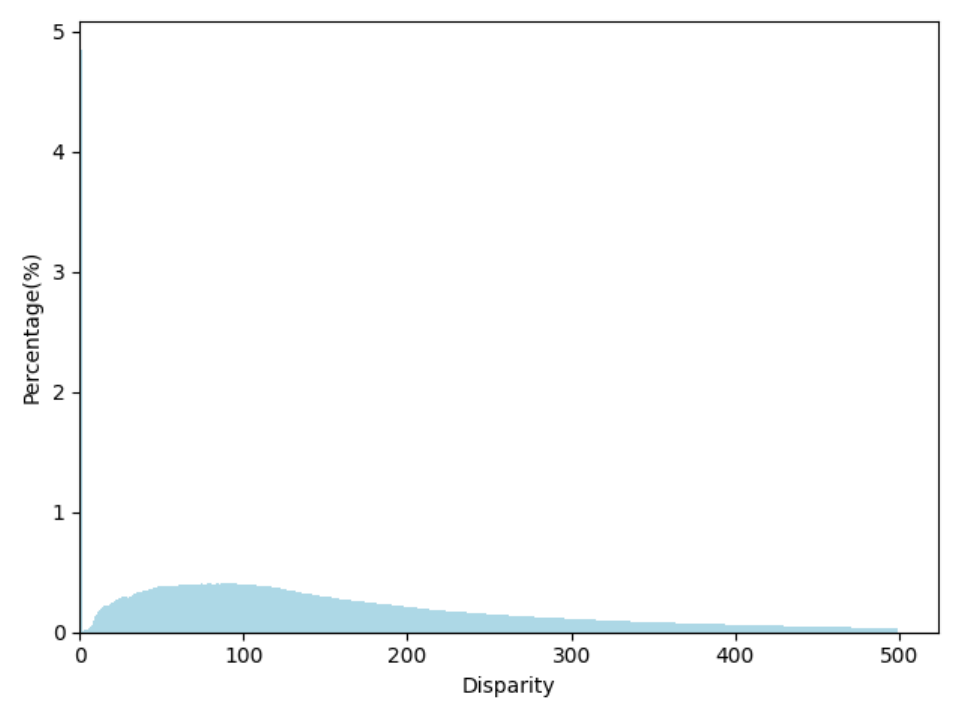}
        \caption{UnrealStereo4k}
    \end{subfigure}
    \begin{subfigure}[b]{0.16\textwidth}
        \centering
        \includegraphics[width=\textwidth]{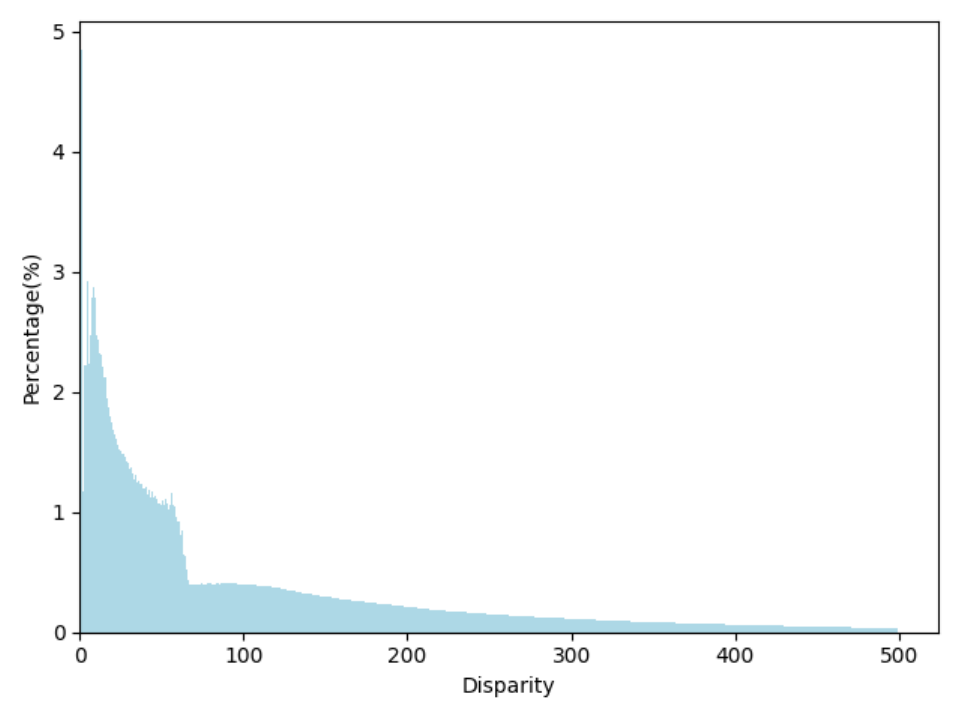}
        \caption{VirtualKITTI2}
    \end{subfigure}
    \begin{subfigure}[b]{0.16\textwidth}
        \centering
        \includegraphics[width=\textwidth]{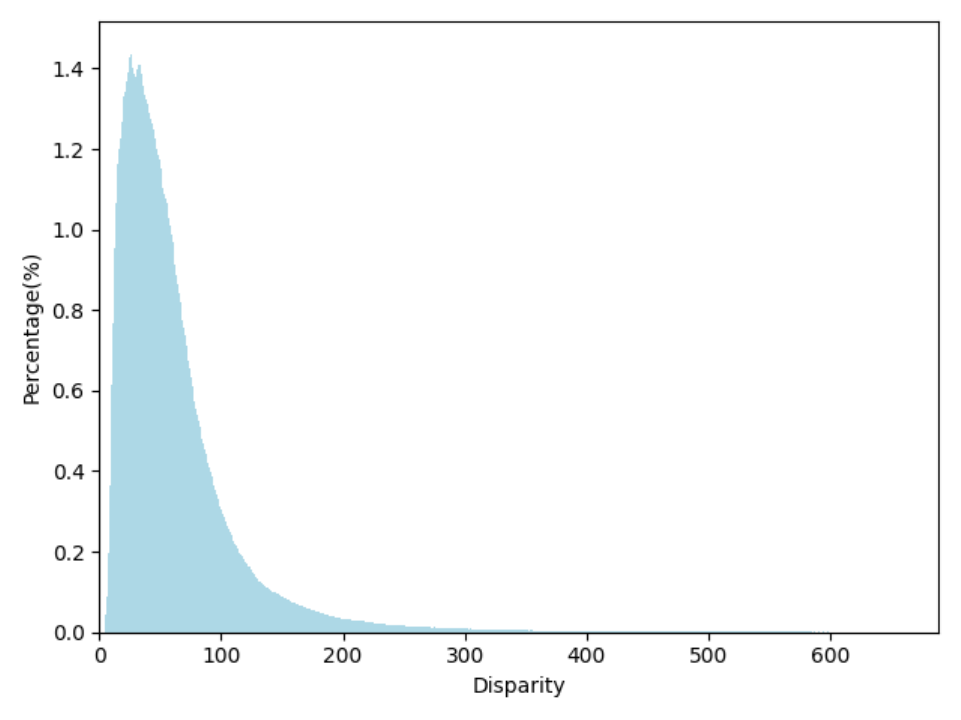}
        \caption{DynamicStereo}
    \end{subfigure}
    \begin{subfigure}[b]{0.16\textwidth}
        \centering
        \includegraphics[width=\textwidth]{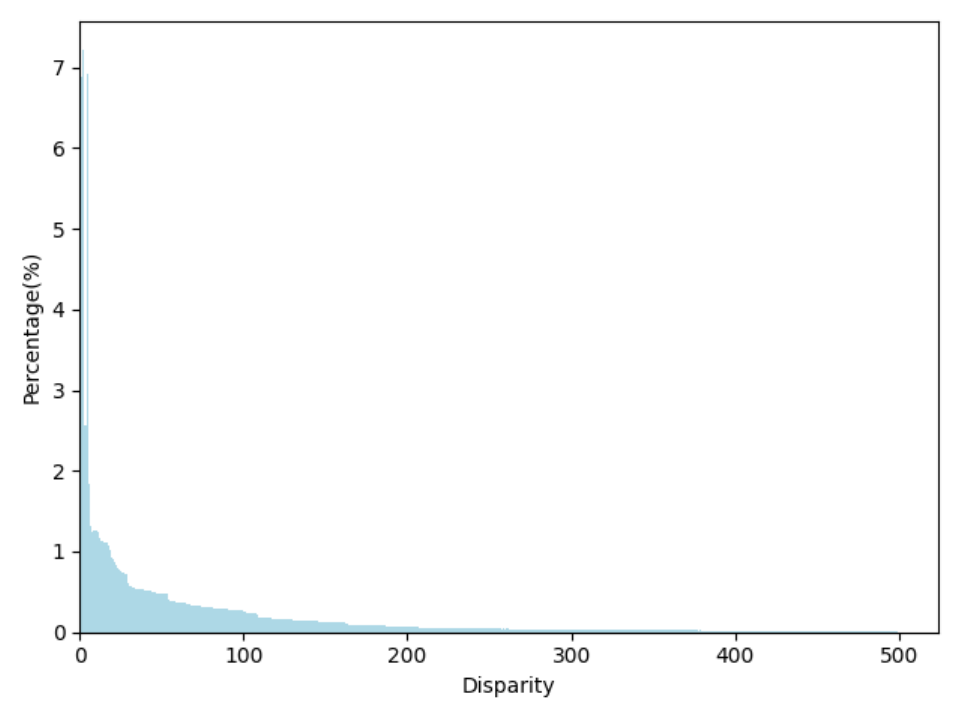}
        \caption{StereoCarla}
    \end{subfigure}
    \begin{subfigure}[b]{0.16\textwidth}
        \centering
        \includegraphics[width=\textwidth]{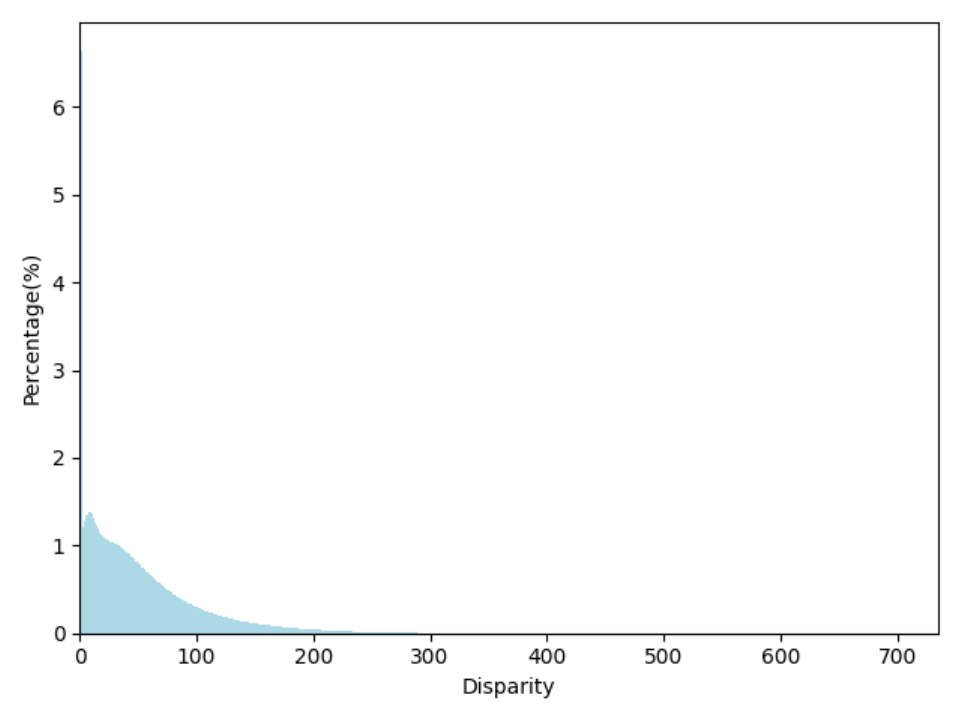}
        \caption{FoundationStereo}
    \end{subfigure}
    
    \begin{subfigure}[b]{0.16\textwidth}
        \centering
        \includegraphics[width=\textwidth]{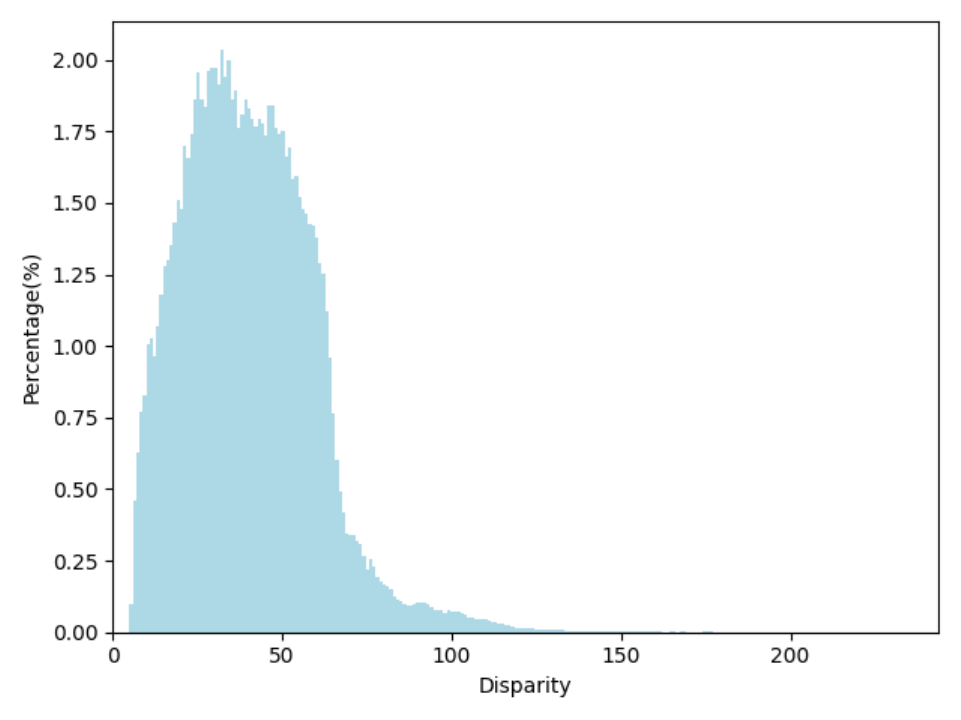}
        \caption{KITTI 2012}
    \end{subfigure}
    \begin{subfigure}[b]{0.16\textwidth}
        \centering
        \includegraphics[width=\textwidth]{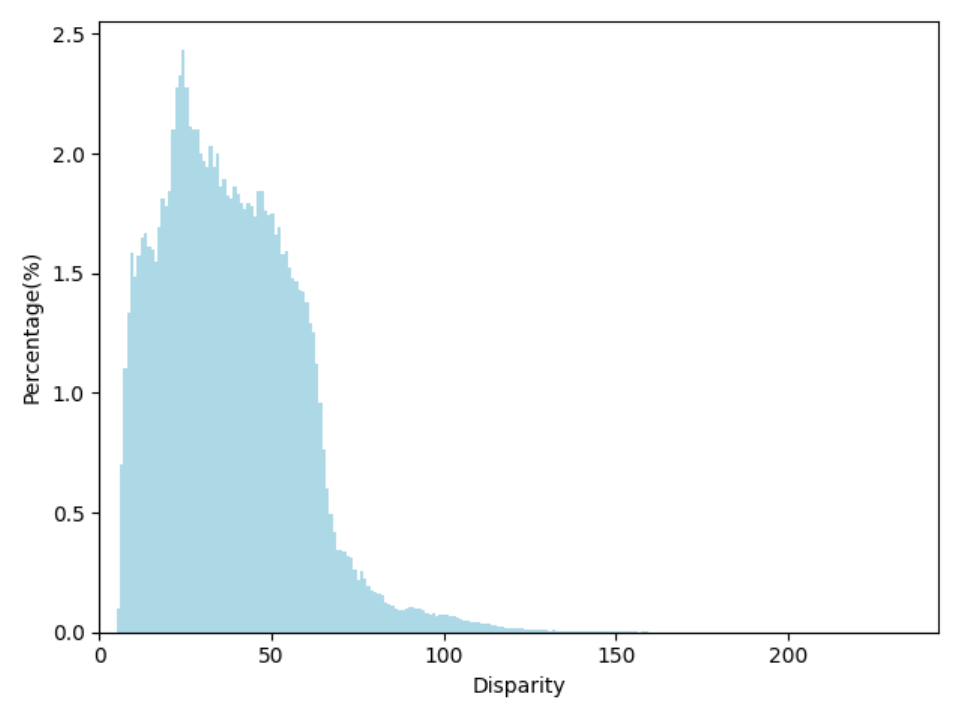}
        \caption{KITTI 2015}
    \end{subfigure}
    \begin{subfigure}[b]{0.16\textwidth}
        \centering
        \includegraphics[width=\textwidth]{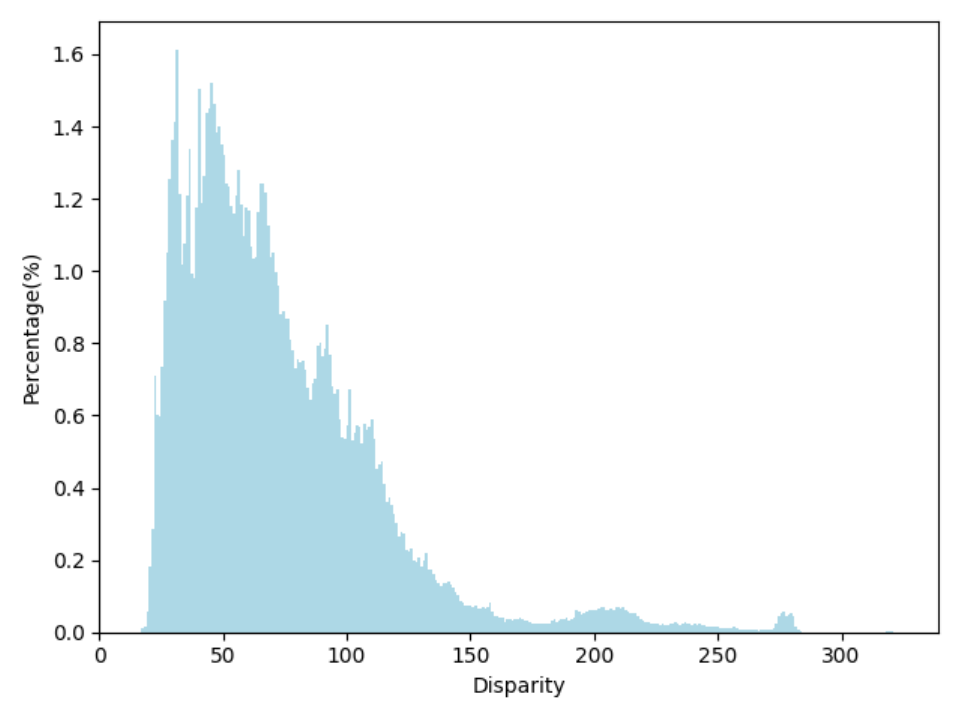}
        \caption{Middlebury}
    \end{subfigure}
    \begin{subfigure}[b]{0.16\textwidth}
        \centering
        \includegraphics[width=\textwidth]{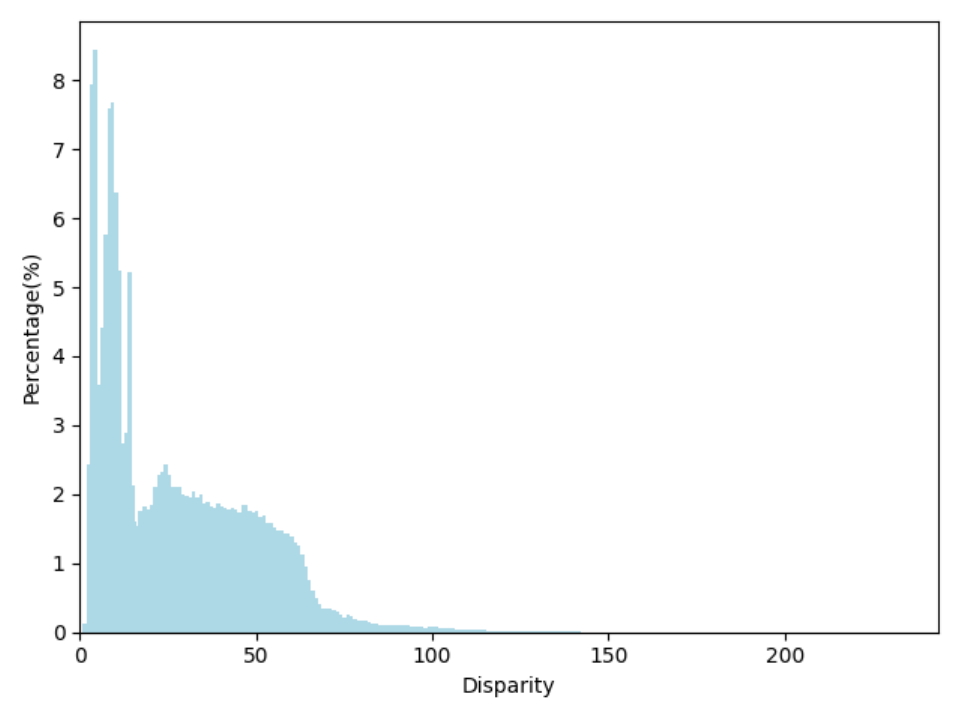}
        \caption{ETH3D}
    \end{subfigure}
    \caption{\textbf{Disparity distribution across different datasets.} We plot the percentage of pixels at different disparities over single datasets. The histograms reveal substantial variations in disparity range, density, and distribution patterns across synthetic and real-world collections, covering indoor and outdoor scenes as well as static and dynamic environments. 
 }  
    \label{fig:distributions}
\end{figure*}

\subsection{Revisiting Labeled Stereo Datasets}
We start by revising existing labeled stereo datasets and unlabeled monocular datasets. Table~\ref{tab:datasets_summary} summarizes the existing labeled datasets involved in our work, dividing them into training and testing sets, along with their properties.

Fig.~\ref{fig:distributions} presents the disparity distribution for all datasets utilized in this work, while Table~\ref{tab:datasets_summary} provides a detailed summary of their acquisition conditions, scene types, and statistical properties. The collections span a broad spectrum of scenarios, encompassing both synthetic and real-world imagery, indoor and outdoor environments, as well as static and dynamic scenes. We also observe substantial variation in acquisition setups, including focal lengths, image resolutions, and baseline distances, which directly influence the disparity range and distribution patterns.

From the distributions, it is evident that certain datasets, such as KITTI~\cite{kitti2012,kitti2015} and Middlebury~\cite{middlebury}, exhibit narrow disparity ranges with peaks concentrated in specific depth intervals—reflecting their targeted acquisition settings and limited environmental diversity. In contrast, datasets like SceneFlow~\cite{sceneflow} and FallingThings~\cite{tremblay2018falling} display significantly broader and more uniform disparity coverage, offering richer supervision signals across multiple depth scales. Synthetic datasets (e.g., UnrealStereo4k, VirtualKITTI2) often provide wide disparity coverage and controlled variations in illumination, weather, and object motion, while real-world LiDAR-based datasets capture fine geometric details but tend to be biased toward certain scene layouts or depth statistics.

This heterogeneity in disparity statistics is crucial for mitigating dataset bias, as models trained solely on narrow-range datasets tend to overfit to their specific depth priors and fail to generalize to unseen environments. By systematically integrating datasets with complementary disparity profiles and scene variations—including high-resolution indoor captures, large-baseline outdoor sequences, and synthetic data with extreme disparity coverage—our mixed-data training corpus fosters the learning of domain-invariant matching features. This diverse foundation enables our Stereo Anything framework to maintain robust zero-shot performance across a wide range of application scenarios.

We also incorporate a large-scale collection of 53.01 million unlabeled monocular images. A comprehensive overview of these datasets is provided in Table~\ref{tab:monocombined}, which are used in the DepthAnything~\cite{depthanytingv1}. These datasets are used to generate synthetic stereo pairs during training.

\subsection{Learning on Labeled Stereo Images}
In this section, we describe our approach to enhancing stereo model generalization by leveraging supervised stereo data. 
Monocular depth models like MiDaS~\cite{midas}, DepthAnytingV1~\cite{depthanytingv1}, and DepthAnytingV2~\cite{depthanytingv2} normalize the depth values to a range of 0 to 1 on each depth map. However, this normalization is clearly unsuitable for stereo matching, as stereo matching has inherent scale information that would be lost during the process. 

Our method focuses on combining multiple existing labeled stereo datasets to create a more diverse and representative training set. 
The key idea is to merge different stereo datasets, each containing varied environmental conditions, scene structures, and disparity distributions, to provide a comprehensive set of training samples. By jointly training on these datasets, our model is exposed to a broader range of disparities and scenes, which helps mitigate overfitting to a single dataset and improves cross-domain performance. The ultimate goal is to create a robust stereo model that can generalize effectively across various real-world scenarios, benefiting from the strengths of each dataset.

Our method begins by training the stereo model on a single dataset and subsequently evaluating its generalization performance across test datasets. To quantify the generalization ability, we calculate the mean metric across these four datasets, producing a ranking of datasets that allows us to assess their impact on cross-dataset performance as $P_{r_1} = f_{eval}(M, D_{1})$. 
Next, we rank all datasets based on evaluation metrics $\{D_{r_1}, D_{r_2}, \dots, D_{r_n}\}$, such that $P_{r_1} \ge P_{r_2} \ge \dots \ge P_{r_n}$ -- i.e., the datasets are ranked from highest to lowest performance.

\begin{figure*}[t!]
    \centering
        \includegraphics[width=\textwidth]{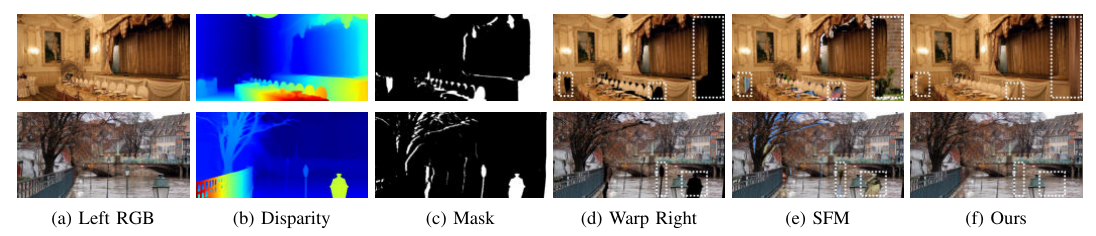}
    \caption{\textbf{Comparison of different methods for filling occlusions in the warped right RGB image.}}  
    \label{fig:realfillCompare}
\end{figure*}

\begin{figure*}[t!]
    \centering
    \includegraphics[width=\textwidth]{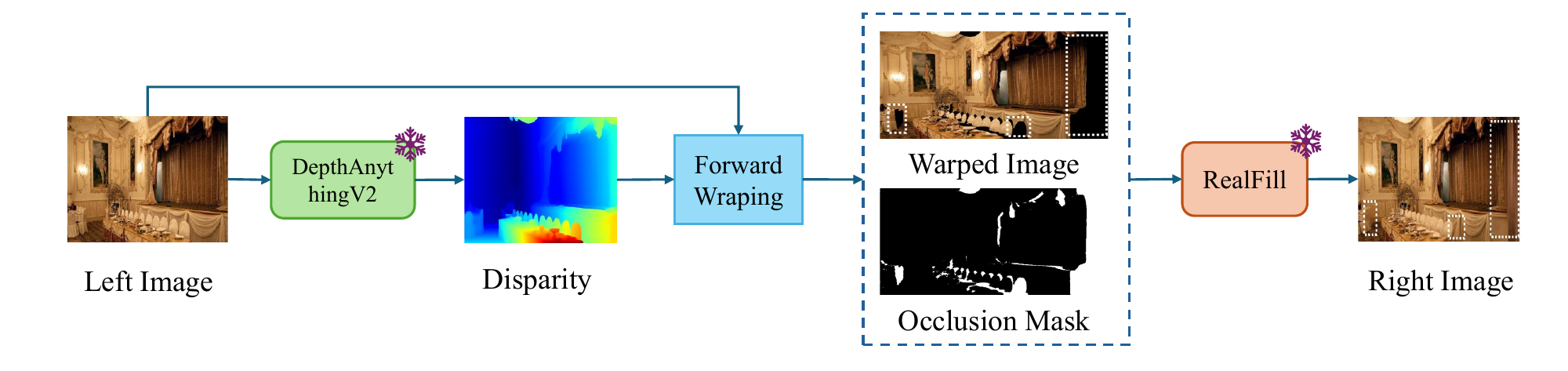}
    \caption{\textbf{Pipeline of our monocular-to-stereo generation.} Given a left image, we first apply DepthAnythingV2 to estimate dense depth, which is then converted into disparity maps. A forward warping process synthesizes a preliminary right view along with a visibility mask highlighting disp-occluded regions. Finally, RealFill~\cite{tang2024realfill} is employed to inpaint the missing areas, producing a high-quality pseudo right image that forms a stereo pair with the original left image.}  
    \label{fig:warpright}
\end{figure*}

The mixing process starts with the top-ranked dataset $\mathcal{D}_{r_1}$, and we sequentially add other datasets to the mixed training set based on their ranking as ${MIX}_k = \bigcup_{i=1}^{k} D_{r_i}$, with $k \in \{1, 2, \dots, n\}$, ensuring that the highest-quality data is prioritized during training.

\subsection{Unleashing the Power of Unlabeled Images}
\label{subsec:unlab}

Although in the previous subsection we collected publicly available labeled stereo data, this data is still insufficient for training a robust foundational model. Additionally, labeled stereo data often requires considerable effort in both data collection and annotation, making it a resource-intensive process.
Works like MiDaS~\cite{midas} and DepthAnything~\cite{depthanytingv1,depthanytingv2} have demonstrated the value of unlabeled images in enhancing data coverage, and this has become a critical component in improving the generalization ability of depth estimation models. However, when it comes to stereo matching, there has been limited exploration into utilizing unlabeled images for scaling up the dataset. Now, it is time for stereo matching to explore data scaling up by taking advantage of the vast resources of unlabeled images available.
One of the primary advantages of this approach is that we can build large-scale, diverse stereo datasets without the need for laborious manual annotation or the computationally intensive process of traditional stereo matching. By leveraging this method, we not only increase the size of our datasets but also ensure their diversity. The unlabeled images can be sourced from a variety of different platforms, containing various scene structures and environmental conditions, thus enriching the dataset in a way that is difficult to achieve with labeled data alone.

Thanks to monocular-based models~\cite{depthanytingv1,depthanytingv2}, stereo-from-mono~\cite{stereo-from-mono}, and image inpainting~\cite{tang2024realfill}, we can effortlessly generate pseudo-stereo datasets from unlabeled monocular images. These generated pseudo-stereo datasets can then be used to train stereo matching models, significantly increasing both the scale and diversity of training data without the traditional manual effort.
The goal is to create a synthesizing stereo image pair and the disparity from single image collections. 
Given a single color image $I$, a pre-trained monocular depth estimation model $g$ is used to estimate the depth $Depth$ of image $I$ as $Depth = g(I)$.
This monocular network can be parameterized as a deep neural network trained either with ground truth depth or through self-supervision.

Next, we need to convert the estimated depth into a disparity map $\tilde{D}$. For simulating realistic stereo pairs with varied baselines and focal lengths, the disparity $\tilde{D}$ is calculated as $\tilde{Disp} = \frac{s\times Depth_{\max}}{Depth}$, where $s$ is a scaling factor randomly sampled from a uniform distribution $[disp_{\min}, disp_{\max}]$, ensuring that the disparities lie within a plausible range.

The goal is to synthesize a new ‘right’ image $\tilde{I}_r$ from the original image $I_l$ and the predicted disparity $\tilde{Disp}$ using forward warping, i.e., by translating each pixel $i$ in $I_l$ by $\tilde{Disp}_i$ to the left. 

During forward warping, certain regions in the right image $\tilde{I}_r$ may remain unfilled due to occlusion, resulting in unnatural gaps, as shown in Figure~\ref{fig:realfillCompare}(d). To mitigate this, previous work~\cite{stereo-from-mono} uses textures from a randomly selected image $I_b$ to fill in the missing pixels, as shown in Figure~\ref{fig:realfillCompare}(e). In contrast, we employ a pre-trained RealFill~\cite{tang2024realfill} model to complete the unfilled regions, ensuring more realistic and context-aware inpainting, as illustrated in Figure~\ref{fig:realfillCompare}(f).

\section{Experiments}

\subsection{Implementation Details}

Our experiments are conducted based on  OpenStereo~\cite{guo2023openstereo}. 
We adopt NMRF-Stereo~\cite{NMRFStereo} as our baseline model due to its strong performance and speed in stereo matching. SwinTransformer~\cite{liu2021swin} is adopted for feature extraction. AdamW optimizer is used for training. For pretraining on SceneFlow~\cite{sceneflow}, we utilize OneCycleLR scheduling with a maximum LR of 0.001, while fine-tuning on other datasets uses OneCycleLR with a maximum LR of 0.0005. 
The batch size is consistently set to 16 across all experiments. Data augmentation techniques, including random crop ($352 \times 640$), color jitter, and random erasing, are applied to enhance the robustness of the model. For fine-tuning on different single training sets, we train the model for a total of 31,250 iterations. When fine-tuning on mixed datasets, we extend the training to 81,000 iterations to guarantee full coverage of all samples within the dataset.
For fine-tuning on pseudo-stereo and combining datasets, we train the model for one epoch due to the large amount of data.
 $[disp_{\min}, disp_{\max}]$ mentioned in \autoref{subsec:unlab} 
are set to [50, 192].

Given the relatively low resolution of ImageNet21K, we resize the images while preserving their original aspect ratio to facilitate training on pseudo-stereo datasets. Specifically, we ensure that the width is at least 768 pixels and the height is at least 384 pixels. This resizing strategy helps preserve the important visual features in the images while maintaining a resolution suitable for effectively training our model. The resolution of the input image is $352 \times 640$.

\subsection{Test datasets and metrics.}
For all experiments, we evaluate our method on four benchmark datasets: KITTI2012~\cite{kitti2012} (194 training image pairs), KITTI2015~\cite{kitti2015} (200 training image pairs), Middlebury~\cite{middlebury} (15 image pairs), and ETH3D~\cite{eth3d} (27 image pairs). Middlebury~\cite{middlebury} is tested on half-resolution. 
Following recent works~\cite{dsmnet2020,xu2023iterative}, we use the D1-all metric for KITTI2012~\cite{kitti2012}, KITTI2015~\cite{kitti2015}, Bad 2.0 for Middlebury~\cite{middlebury}, and Bad 1.0 for ETH3D~\cite{eth3d}.
The D1-all computes the percentage of pixels with an absolute disparity error larger than 3.0 pixels. Bad 1.0/Bad 2.0 metric measures the percentage of pixels with a disparity error greater than 1.0/2.0 pixel. 

\begin{figure*}[t!]
    \centering
    \begin{subfigure}[b]{\textwidth}
         \includegraphics{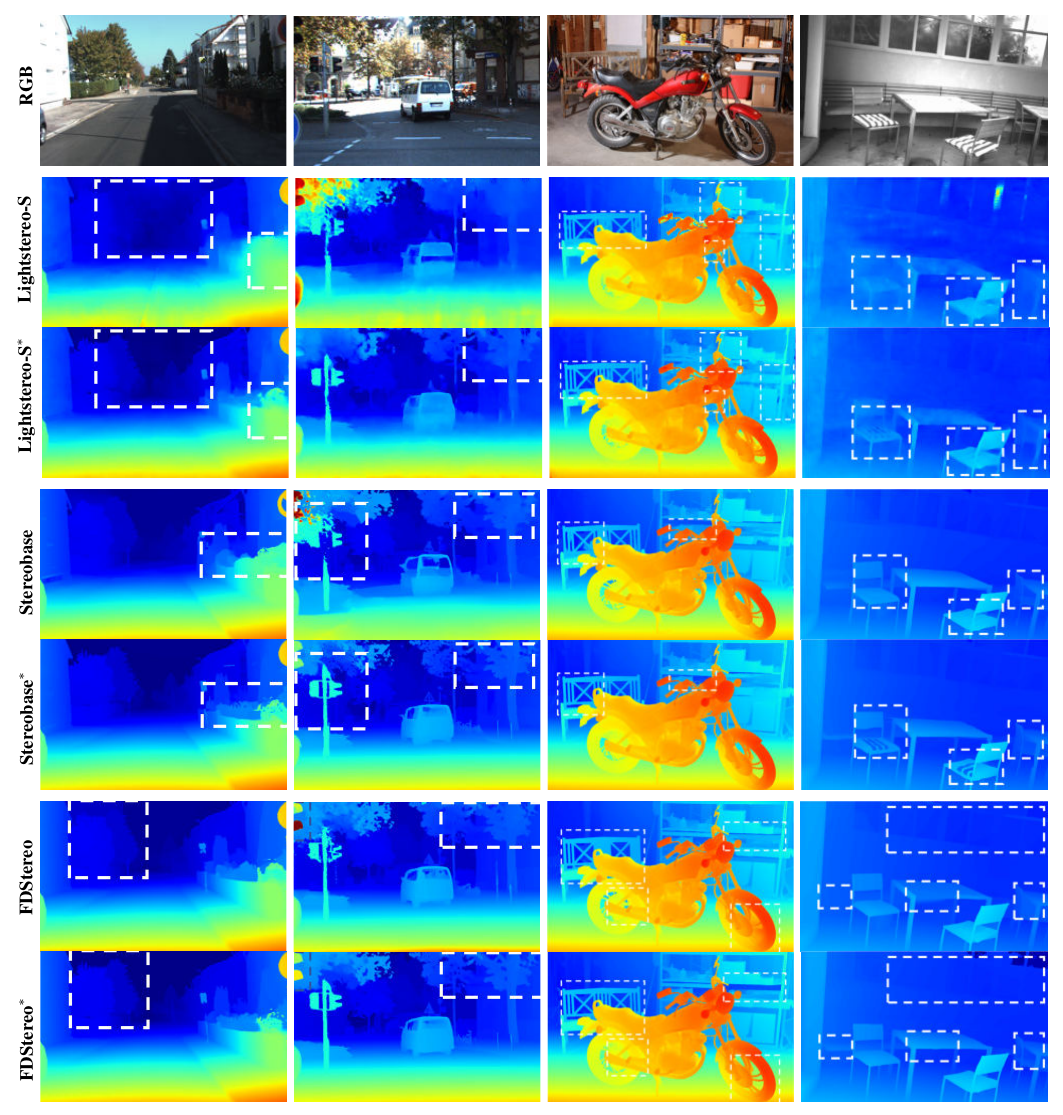} 
    \end{subfigure}
    \caption{Comparison of disparity estimation results from different models on four benchmark datasets. The first column shows the input RGB images, followed by the disparity maps predicted by three representative stereo models: the speed-focused LightStereo-S~\cite{guo2025lightstereo}, the accuracy-focused StereoBase~\cite{guo2023openstereo}, and FoundationStereo~\cite{wen2025foundationstereo}, which incorporates monocular priors from foundation models. Models trained with our proposed strategy are marked with ($^*$). The dashed and solid circles highlight key areas where our models show notable improvements in depth accuracy, especially in textureless and occluded regions. 
}
    \label{fig:stereo_comparison}
\end{figure*}

\begin{table}[t]
\center
\small
\caption{\textbf{Comparison with SOTA methods} including the computed Mean column. The Mean is the average of the four metrics (K12, K15, Midd, E3D). Lower is better. \textbf{Bold}: Best in each row. 
}
\label{tab:compareSOTA}
\centering
\setlength\tabcolsep{5pt}
\renewcommand\arraystretch{1.1}
\begin{tabular}{r||cccc|c}
\thickhline
\rowcolor{gray!20} 
\textbf{Method} &\textbf{K12} & \textbf{K15} & \textbf{Midd} & \textbf{E3D}& \textbf{Mean} \\
\hline\hline
PSMNet~\cite{psmnet2018} & 30.51 & 32.15 & 33.53 & 18.02 & 28.55 \\
\rowcolor{gray!10}
GwcNet~\cite{gwcnet2019} & 23.05 & 25.19 & 29.87 & 14.54 & 23.16 \\
CFNet~\cite{cfnet2021} & 13.64 & 12.09 & 23.91 & 7.67 & 14.33 \\
\rowcolor{gray!10}
COEX~\cite{bangunharcana2021coex} & 12.08 & 11.01 & 25.17 & 11.43 & 14.92 \\
FADNet++~\cite{wang2021fadnet++} & 11.31 & 13.23 & 24.07 & 22.48 & 17.77 \\
\rowcolor{gray!10}
LightStereo-L~\cite{guo2025lightstereo} & 6.41 & 6.40 & 17.51 & 11.33 & 10.41 \\
DSMNet~\cite{dsmnet2020}& 6.2 &6.5 &8.1 &6.2&6.75\\  
\rowcolor{gray!10}
RAFT-Stereo~\cite{raftstereo}& 4.7 &5.5 &9.4 &3.3&5.72\\
IGEV~\cite{xu2023iterative} & 4.88 & 5.16 & 8.47 & 3.53 & 5.51 \\
\rowcolor{gray!10}
Selective-IGEV~\cite{SelectiveStereo}& 5.64 &6.05& 12.04& 5.40&7.28\\ 
StereoBase~\cite{guo2023openstereo} & 4.85 & 5.35 & 9.76 & 3.12 & 5.77 \\
\rowcolor{gray!10}
NMRF~\cite{NMRFStereo} & 4.20 & 5.10 & 7.50 & 3.80 & 5.15 \\
MonSter~\cite{cheng2025monster}& 3.62& 3.97& 5.17 &2.03&3.70\\
\rowcolor{gray!10}
DEFOM~\cite{defomstereo}&3.76 & 4.99&  3.26&  2.35&3.59\\
FoundationStereo~\cite{wen2025foundationstereo} & 3.2&4.9&5.5 &1.8&3.85\\
\hline
\rowcolor{cyan!10}
\textbf{StereoAnything} &\textbf{3.01}&\textbf{3.26}&\textbf{2.69}&\textbf{0.77}&\textbf{2.68}
\\
\bottomrule
\end{tabular}
\end{table}

\begin{table}[t]
\center
\small
\caption{\textbf{Ablation study results in SOTA comparison format.} $^\dagger$ indicates use of our training strategy. $^*$ indicates NMRFStereo-SwinT. Lower is better. \textbf{Bold}: Best overall result.}
\label{tab:ablationSOTAstyle}
\centering
\setlength\tabcolsep{3pt}
\renewcommand\arraystretch{1.1}
\begin{tabular}{r||cccc|c}
\thickhline
\rowcolor{gray!20}
\textbf{Method} & \textbf{K12} & \textbf{K15} & \textbf{Midd} & \textbf{E3D}& \textbf{Mean} \\
\hline\hline
\multicolumn{6}{l}{\textbf{Speed-focused model}} \\ \hline
LightStereo-S~\cite{guo2025lightstereo} & 11.57 & 9.01 & 19.63 & 29.39 & 17.40 \\
\rowcolor{gray!10}
LightStereo-S~\cite{guo2025lightstereo}$^\dagger$&\textbf{4.83}& \textbf{5.06}&\textbf{13.03}& \textbf{7.62}&\textbf{7.64}\\
& \reddown{6.74} & \reddown{3.95} & \reddown{6.60} & \reddown{21.77} & \reddown{9.76} \\
    
\hline
LightStereo-L~\cite{guo2025lightstereo} & 6.41 & 6.40 & 17.51 & 11.33 & 10.41 \\
\rowcolor{gray!10}
LightStereo-L~\cite{guo2025lightstereo}$^\dagger$&\textbf{4.20}& \textbf{4.78}&\textbf{9.46}& \textbf{3.97}&\textbf{5.60}\\
& \reddown{2.21} & \reddown{1.62} & \reddown{8.05} & \reddown{7.36} & \reddown{4.81} \\

\hline
\multicolumn{6}{l}{\textbf{Accuracy-focused model}} \\ \hline
StereoBase~\cite{guo2023openstereo}	&4.85	&5.35	&9.76	&3.12	&5.77\\
\rowcolor{gray!10}
StereoBase~\cite{guo2023openstereo}$^\dagger$&\textbf{3.45}&\textbf{4.43}&\textbf{5.64}&\textbf{2.05}&\textbf{3.89}\\
& \reddown{1.40} & \reddown{0.92} & \reddown{4.12} & \reddown{1.07} & \reddown{1.88} \\

\hline
NMRF~\cite{NMRFStereo} & 4.20 & 5.10 & 7.50 & 3.80 & 5.15 \\
\rowcolor{gray!10}
NMRF~\cite{NMRFStereo}$^\dagger$&\textbf{3.27}& \textbf{3.97}&\textbf{6.73}& \textbf{1.72}&\textbf{3.92}\\
& \reddown{0.93} & \reddown{1.13} & \reddown{0.77} & \reddown{2.08} & \reddown{1.23} \\

\hline
NMRF-SwinT~\cite{NMRFStereo}$^*$&8.67&7.46&16.36&23.46 &13.99\\
\rowcolor{gray!10}
NMRF-SwinT~\cite{NMRFStereo}$^*\dagger$ &\textbf{3.48}&\textbf{3.88}&\textbf{7.03}&\textbf{1.83}&\textbf{4.06}\\
& \reddown{5.59} & \reddown{3.58} & \reddown{9.33} & \reddown{21.63} & \reddown{9.93} \\
\hline

FoundationStereo~\cite{wen2025foundationstereo} & 3.2&4.9&5.5 &1.8&3.85\\
\rowcolor{gray!10}
FoundationStereo~\cite{wen2025foundationstereo}$^\dagger$ &\textbf{3.01}&\textbf{3.26}&\textbf{2.69}&\textbf{0.77}&\textbf{2.68}\\
& \reddown{0.19} & \reddown{1.64} & \reddown{2.81} & \reddown{1.03} & \reddown{1.17} \\

\bottomrule
\end{tabular}
\end{table}

\begin{table*}[t] 
\small
\centering
\caption{\textbf{Combinations of labeled stereo datasets.}}%
  \setlength\tabcolsep{5pt}
  \renewcommand\arraystretch{1.1}
\begin{tabular}{l|ccccccccccc||cccc|c}
  \thickhline
  \rowcolor{gray!20}
\textbf{Mix}   & \textbf{FSD}&\textbf{SC} & \textbf{Tar}&  \textbf{CRE} & \textbf{SP}& \textbf{ST} &\textbf{DR}& \textbf{FT}&\textbf{I2K} &\textbf{VK2} &\textbf{U4K}& \textbf{K12} & \textbf{K15} & \textbf{Midd} & \textbf{E3D} & \textbf{Mean}  \\ %
  \hline\hline

  MIX 1 &\checkmark &&&&& &         &          &     &&&4.45&	3.67&	6.59	&3.20&	4.48  \\
  \rowcolor{gray!10}
  MIX 2 &\checkmark & \checkmark & &&&&     &&  &&&3.58&	3.62&	6.58	&2.29&	4.02\\
  MIX 3 &\checkmark & \checkmark &   \checkmark   & & &&&&&  &&3.6&	3.66&	6.54	&2.27&	4.02\\  %
  \rowcolor{gray!10}
  MIX 4 &\checkmark & \checkmark & \checkmark & \checkmark        &&&&&&&& 3.67&	3.77&	6.43&	2.12&	4.00\\
  MIX 5 & \checkmark & \checkmark &  \checkmark&\checkmark & \checkmark &&&&& &&3.62	&3.77	&6.35&	1.98	&3.93\\
  \rowcolor{gray!10}
  MIX 6 & \checkmark & \checkmark & \checkmark &\checkmark & \checkmark& \checkmark &&&&&&3.63&	3.77&	6.55&	2.32&	4.07 \\
  MIX 7 & \checkmark & \checkmark & \checkmark &\checkmark & \checkmark & \checkmark& \checkmark&&&&&3.59	&3.76&	6.21&	\textbf{1.95}	&3.88 \\
  \rowcolor{gray!10}
  MIX 8 & \checkmark & \checkmark & \checkmark &\checkmark & \checkmark & \checkmark& \checkmark& \checkmark& && &3.55&	3.75&	6.42&	1.98	&3.93 \\
  MIX 9 & \checkmark & \checkmark & \checkmark &\checkmark & \checkmark& \checkmark& \checkmark& \checkmark& \checkmark&&&3.59&	3.75	&6.33	&2.13	&3.95 \\%
  \rowcolor{gray!10}
  MIX 10 & \checkmark & \checkmark & \checkmark &\checkmark & \checkmark& \checkmark& \checkmark& \checkmark& \checkmark&\checkmark& &\textbf{3.49}	&\textbf{3.56}	&6.18	&2.10&	\textbf{3.83} \\%
  MIX 11 & \checkmark & \checkmark & \checkmark &\checkmark & \checkmark& \checkmark& \checkmark& \checkmark& \checkmark&\checkmark&\checkmark&3.54	&3.63&	6.25	&1.98&	3.85 \\%
  \thickhline
  
\end{tabular}
\label{tab:dataset_mix}
\end{table*}
\begin{table}[t]
  \small
  \centering
  \caption{\textbf{\textbf{Cross-domain evaluation} on different labeled stereo datasets.} Relative to baseline (top row), \textcolor{OliveGreen}{green}/\textcolor{red}{red} shows performance improvement/decline. $^*$ indicates the results from the StereoCarla~\cite{stereocarla} paper. \textbf{Bold}: Best. \underline{Underline}: Second Best.}
  \setlength\tabcolsep{2pt}
  \renewcommand\arraystretch{1.1}
\begin{tabular}{l||cccc|cc}
\thickhline
\rowcolor{gray!20}
\textbf{Dataset} & \textbf{K12} & \textbf{K15} & \textbf{Midd} & \textbf{E3D} & \textbf{Mean} & \textbf{Rank} \\
\hline\hline
SF & 8.67 & 7.46 & 16.36 & 23.46 & 13.99 & - \\
\hline
\rowcolor{gray!10} 
SF$\rightarrow$FoundationStereo & \textcolor{OliveGreen}{4.13}	&	\textbf{\textcolor{OliveGreen}{3.64}}&	\textbf{\textcolor{OliveGreen}{6.56}}&	\textbf{\textcolor{OliveGreen}{2.76}}	&\textbf{4.27} & 1 \\

SF$\rightarrow$StereoCarla$^*$  & \underline{\textcolor{OliveGreen}{4.11}} & \textcolor{OliveGreen}{4.87} & \underline{\textcolor{OliveGreen}{9.12}} & \underline{\textcolor{OliveGreen}{3.17}} & \underline{5.32} & 2 \\

\rowcolor{gray!10} 
SF$\rightarrow$Tartanair$^*$ & \textcolor{OliveGreen}{4.16} & \textcolor{OliveGreen}{4.71} & \textcolor{OliveGreen}{13.95} & \textcolor{OliveGreen}{5.25} & 7.02 & 3 \\

SF$\rightarrow$CREStereo$^*$ & \textcolor{OliveGreen}{8.01} & \textcolor{OliveGreen}{6.18} & \textcolor{OliveGreen}{13.73} & \textcolor{OliveGreen}{5.75} & 8.42 & 4 \\

\rowcolor{gray!10} 
SF$\rightarrow$Spring$^*$ & \textcolor{OliveGreen}{6.59} & \textcolor{OliveGreen}{6.23} & \textcolor{OliveGreen}{16.04} & \textcolor{OliveGreen}{6.96} & 8.96 & 5 \\

SF$\rightarrow$Sintel$^*$& \textcolor{OliveGreen}{6.09} & \textcolor{OliveGreen}{6.28} & \textcolor{red}{19.28} & \textcolor{OliveGreen}{6.18} & 9.46 & 6 \\

\rowcolor{gray!10} 
SF$\rightarrow$DynamicReplica$^*$ & \textcolor{red}{11.84} & \textcolor{red}{15.36} & \textcolor{OliveGreen}{12.84} & \textcolor{OliveGreen}{5.32} & 11.34 & 7 \\

SF$\rightarrow$FallingThings$^*$ & \textcolor{OliveGreen}{4.28} & \textcolor{OliveGreen}{4.23} & \textcolor{OliveGreen}{13.17} & \textcolor{red}{27.93} & 12.40 & 8 \\

\rowcolor{gray!10} 
SF$\rightarrow$Instereo2K$^*$ & \textcolor{red}{13.33} & \textcolor{red}{15.21} & \textcolor{OliveGreen}{11.75} & \textcolor{OliveGreen}{11.23} & 12.88 & 9 \\

SF$\rightarrow$VirtualKitti2$^*$ & \textbf{\textcolor{OliveGreen}{3.96}} & \underline{{\textcolor{OliveGreen}{4.00}}} & \textcolor{red}{22.23} & \textcolor{red}{73.78} & 25.99 & 10 \\

\rowcolor{gray!10} 
SF$\rightarrow$UnrealStereo4K$^*$ & \textcolor{red}{8.68} & \textcolor{OliveGreen}{6.90} & \textcolor{red}{44.98} & \textcolor{red}{64.51} & 31.27 & 11 \\


\thickhline

\end{tabular}

\label{tab:onetraincolor}
\end{table}

\subsection{Comparison with other SOTA methods}
As shown in \ref{tab:compareSOTA}, we compare our best-performing model with several SOTA stereo-matching methods.  
Our model, StereoAnything, consistently outperforms all the compared methods in terms of the mean error across all four datasets, achieving a significant reduction in the overall performance. Specifically, StereoAnything achieves a mean error of \textbf{2.68}, which is considerably lower than the best-performing method from previous works. In particular, StereoAnything significantly reduces the error on challenging datasets such as ETH3D, where the disparity is often more complex due to the presence of textureless regions and occlusions. 
Overall, these results highlight the effectiveness of our hybrid training strategy, combining labeled stereo datasets and pseudo-stereo data generated from monocular depth estimation models. The superior performance of StereoAnything across multiple datasets emphasizes its strong generalization capabilities, making it a highly competitive solution for stereo matching tasks in both benchmark and real-world scenarios.

\subsection{Effectiveness of Our Training Strategy}
To validate the effectiveness of our proposed training strategy, we conduct both quantitative and qualitative analyses across a variety of stereo models and benchmarks. 

Table~\ref{tab:ablationSOTAstyle} presents the ablation study in a SOTA comparison format, where we apply our training strategy (denoted by $\dagger$) to several representative stereo models of different capacities. Across all benchmarks, we observe consistent and significant improvements, confirming the universality of our approach. For lightweight architectures such as LightStereo-S and LightStereo-L, our strategy leads to dramatic performance gains: on ETH3D, LightStereo-S reduces the error from 29.39 to 7.62 (a 21.77 point drop), while LightStereo-L decreases the mean error from 10.41 to 5.60. These results indicate that our large-scale mixed-data training is particularly effective in enhancing the generalization ability of compact models. More powerful baselines such as StereoBase and NMRFStereo also benefit substantially, achieving error reductions of up to 1.88 and 1.23 points on the mean metric, respectively. Notably, even the recent FoundationStereo—already designed with strong generalization in mind—further improves from 3.85 to 2.84 in the mean metric, suggesting that our training strategy provides complementary gains beyond architectural design. 
Overall, these results highlight two key advantages of our method: (1) it scales effectively across models of varying capacities, from lightweight designs to foundation-level architectures; and (2) it yields especially large improvements on challenging benchmarks such as Middlebury and ETH3D, demonstrating its robustness and cross-domain generalization capability.

\textbf{Qualitative Results.} Fig.~\ref{fig:stereo_comparison} further illustrates these improvements qualitatively. With our proposed training strategy, LightStereo-S generates sharper edges and better preserves thin structures that are often blurred in the baseline predictions. StereoBase demonstrates cleaner object boundaries and smoother depth transitions, particularly in complex scenes. These visual comparisons confirm that our approach consistently enhances disparity quality across different models and datasets. Both the quantitative and qualitative results underscore the robustness and universality of our training strategy. It effectively scales from lightweight to foundation-level architectures, delivers consistent improvements across diverse benchmarks, and significantly enhances cross-domain generalization.

\subsection{Results on labeled stereo datasets}
In this section, we are going to ablate our StereoAnything framework, showing in detail how different datasets impact the final results.

\subsubsection{Training on labeled stereo datasets.} 

\textbf{Training on single-labeled stereo datasets.} 
To assess the impact of different training datasets on the generalization capabilities of the stereo-matching model, we conducted extensive experiments. We fine-tuned the pre-trained model on SceneFlow~\cite{sceneflow} to various synthetic and real-world datasets.
As shown in Table~\ref{tab:onetraincolor}, fine-tuning on synthetic datasets such as Sintel~\cite{Sintel}, FallingThings~\cite{tremblay2018falling}, and Tartanair~\cite{wang2020tartanair} generally leads to improved performance over the baseline. 
However, fine-tuning on certain datasets like  UnrealStereo4K~\cite{tosi2021smd} results in significantly degraded performance, with mean metrics increasing to 50.77\% and 31.27\%, respectively. This suggests that these datasets may not provide sufficient diversity or introduce biases hindering the model's generalization of different scenarios. 
In summary, the choice of fine-tuning dataset plays a crucial role in the generalization ability of stereo-matching models. Datasets that closely mimic the diversity and complexity of real-world environments significantly enhance performance across multiple benchmarks.

\begin{table*}[t]
\small
    \centering
    \caption{\textbf{Ablation study.} pseudo-stereo generated from different monocular datasets. $*$ indicates that the original image resolution is smaller than $352 \times 640$. In such cases, the original RGB image is resized before being processed by the depth model, after which the pseudo-stereo pairs are synthesized. M refers to a million images.}
    \setlength\tabcolsep{5pt}
    \renewcommand\arraystretch{1.1}
    \begin{tabular}{l|ccc||cccc|c}
        \thickhline
        \rowcolor{gray!20}
        \textbf{Dataset} & \textbf{Indoor} & \textbf{Outdoor} &  \textbf{Images} & \textbf{Kitti12} & \textbf{Kitti15} & \textbf{Middlebury} & \textbf{Eth3D} & \textbf{Mean} \\
        \hline\hline
        LSUN~\cite{yu2015lsun}$*$ & \checkmark & & 9.8M & 4.27 & 4.17 & 11.88 & 3.91 & 6.06 \\
        \rowcolor{gray!10}
        ImageNet-21K~\cite{imagenet21k}$*$ & \checkmark & & 13.1M & 4.10 & 4.43 & 10.44 & 4.25 & 5.81 \\
        BDD100K~\cite{bdd100k} & & \checkmark & 0.1M & \textbf{3.69} & \textbf{3.87} & 11.75 & 3.92 & 5.81 \\
        \rowcolor{gray!10}
        Google Landmarks~\cite{weyand2020GLDv2} & & \checkmark & 4.97M & 3.91 & \underline{4.01} & \underline{10.31} & \underline{3.36} & \underline{5.40} \\
        Places365~\cite{zhou2017places} & \checkmark & \checkmark & 10.2M & 3.98 & 4.16 & 10.68 & 3.73 & 5.64 \\
        \rowcolor{gray!10}
        Open Images V7~\cite{kuznetsova2020open} & \checkmark & \checkmark & 1.9M & 4.15 & 4.25 & 11.27 & 3.75 & 5.86 \\
        SA-1B~\cite{kirillov2023segment} & \checkmark & \checkmark & 11.1M & 4.24 & 4.89 & 12.32 & 4.41 & 6.47 \\
        \rowcolor{gray!10}
        Objects365~\cite{shao2019objects365}$*$ & \checkmark & \checkmark & 2.0M & \underline{3.78} & 4.12 & \textbf{9.79} & \textbf{3.33} & \textbf{5.26} \\
        ALL & \checkmark & \checkmark & 53.17M & 4.22 & 4.69 & 12.96 & 4.99 & 6.72 \\
        \thickhline
    \end{tabular}
    \vspace{0.5cm}
    \label{tab:monocombined}
\end{table*}

\begin{table}[t]
\small
    \centering
    \caption{\textbf{Ablation study.} different monocular depth estimation methods (Top) and RGB completion (Bottom) on the BDD100K~\cite{bdd100k}. Random indicates that a randomly selected RGB image fills the masked regions in the warped right RGB image.}
    \setlength\tabcolsep{3pt}
    \renewcommand\arraystretch{1.1}
    \begin{tabular}{lc||cccc|c}
        \thickhline
        \rowcolor{gray!20}
        \textbf{Method} & \textbf{Completion}&\textbf{K12} & \textbf{K15} & \textbf{Midd} & \textbf{E3D} & \textbf{Mean}\\
    \hline\hline
     MiDas~\cite{midas}& Random & 4.21 & 4.23 & 12.04 & 4.70 & 6.30\\
     \rowcolor{gray!10}
    DAv1~\cite{depthanytingv1}& Random& 4.01 & 3.98 & 12.08 & 4.48 & 6.14\\
    DAv2~\cite{depthanytingv2}& Random & \textbf{3.69} & \textbf{3.87} & \textbf{11.75} & \textbf{3.92} & 5.81 \\
    \midrule
     \rowcolor{gray!10}
    DAv2~\cite{depthanytingv2}& Random & 3.69 & \textbf{3.87} & 11.75 & \textbf{3.92} & 5.81 \\
    DAv2~\cite{depthanytingv2} &RealFill&\textbf{3.68}&3.88&\textbf{9.39}&3.93&\textbf{5.22}\\
        \thickhline
    \end{tabular}
    \label{tab:monoCompletion}
\end{table}

\textbf{Training on mixed labeled stereo datasets.} 
In the mixed experiments, we begin with FoundationStereo dataset~\cite{wen2025foundationstereo}, as it achieved the best performance in the ~Table~\ref{tab:onetraincolor}. Then, similar to~\cite{midas}, we sequentially add other datasets to the mixed dataset according to the ranking results in Table~\ref{tab:onetraincolor}.
Table~\ref{tab:dataset_mix} shows the detailed performance results across these benchmarks for each mix configuration. Among the tested configurations, MIX 10 achieves the best overall performance with the lowest mean error (3.83), demonstrating superior generalization across all datasets.
Notably, it achieves the best results on K12(3.49) and K15 (3.56). These findings highlight the importance of dataset selection and mixing strategies in stereo matching

\subsection{Results on large-scale unlabeled data}
We also incorporate a large-scale collection of 53.17 million unlabeled monocular images. A comprehensive overview of these datasets \cite{depthanytingv1} is provided in Table~\ref{tab:monocombined}. These datasets are used to generate synthetic stereo pairs during training.

\textbf{Effectiveness of monocular depth method.} In Table~\ref{tab:monoCompletion} (top), we validate the impact of using different monocular depth estimation models to generate pseudo stereo data on the BDD100k~\cite{bdd100k} dataset. Fine-tuning on the stereo datasets synthesized using MiDas~\cite{midas}, DepthAnytingV1~\cite{depthanytingv1} (DAv1), and DepthAnytingV2~\cite{depthanytingv2} (DAv2) leads to competitive results, demonstrating the effectiveness of using monocular depth models for pseudo-stereo data generation. Specifically, the synthesized data from DAv2~\cite{depthanytingv2} outperforms other monocular methods across all benchmarks. Therefore, DepthAnytingV2~\cite{depthanytingv2} is used for the subsequent experiments.

\textbf{Effectiveness of monocular datasets.} We evaluate the impact of using different monocular datasets on the performance of the pseudo stereo data generation, based on the DAv2~\cite{depthanytingv2} model.
As shown in Table~\ref{tab:monocombined}, the Objects365~\cite{shao2019objects365} achieves the best performance overall, with a mean metric of 5.26, benefiting from its balanced coverage of both indoor and outdoor scenes as well as higher-quality imagery.  
In contrast, datasets such as LSUN, ImageNet-21K, and Objects365 must be resized before depth estimation and pseudo-stereo synthesis because their original resolution is smaller than $352 \times 640$. This resizing inevitably introduces artifacts and degrades the quality of the generated pairs, thereby limiting the final stereo performance.
Interestingly, the combined setting (ALL) does not yield the best results despite leveraging the largest data volume (over 53M images). We attribute this to domain conflicts and noise accumulation: merging heterogeneous datasets with different scene distributions, resolutions, and levels of annotation quality can dilute the beneficial signals from high-quality subsets. This highlights that simply scaling the quantity of pseudo-stereo data is insufficient; instead, the choice and quality of source datasets play a more critical role. 
Therefore, we select Objects365~\cite{shao2019objects365} as the primary dataset for subsequent experiments.

\textbf{Effectiveness of image completion.}
Finally, we evaluate the impact of different image completion methods on stereo matching performance in Table~\ref{tab:monoCompletion} (bottom). Compared to the Stereo-from-Mono (SFM) approach~\cite{stereo-from-mono}, which uses random RGB images for filling in missing pixels, our method achieves better results across all datasets. Specifically, the mean error decreases from 5.81 with SFM to 5.22, demonstrating the effectiveness of our image completion strategy. 

\begin{table}[t] 
\small
    \centering
     \caption{\textbf{Cross-domain evaluation combines mixed labeled stereo and pseudo-stereo datasets.}}
    \setlength\tabcolsep{5pt}
    \renewcommand\arraystretch{1.1}
    \begin{tabular}{l||cccc|c}
    \thickhline
    \rowcolor{gray!20}
    \textbf{Dataset}& \textbf{K12} & \textbf{K15} & \textbf{Midd} & \textbf{E3D} & \textbf{Mean}   \\
    \hline\hline
     MIX 10  &\textbf{3.49}	&3.56	&\textbf{6.18}	&2.10&	\textbf{3.83} \\
     \rowcolor{gray!10} 
     Obj~\cite{shao2019objects365}&3.78 & 4.12 & 9.79 & 3.33 & 5.26\\
     \midrule
    MIX10+Obj&\textbf{3.49}&\textbf{3.53}&6.98&\textbf{1.85}&3.96 \\
    \thickhline
    \end{tabular}
    \label{tab:combine}
\end{table}

\subsection{Training on labeled and pseudo stereo datasets.} 

\begin{figure*}[t]  
    \centering 
    \includegraphics[width=0.98\textwidth]{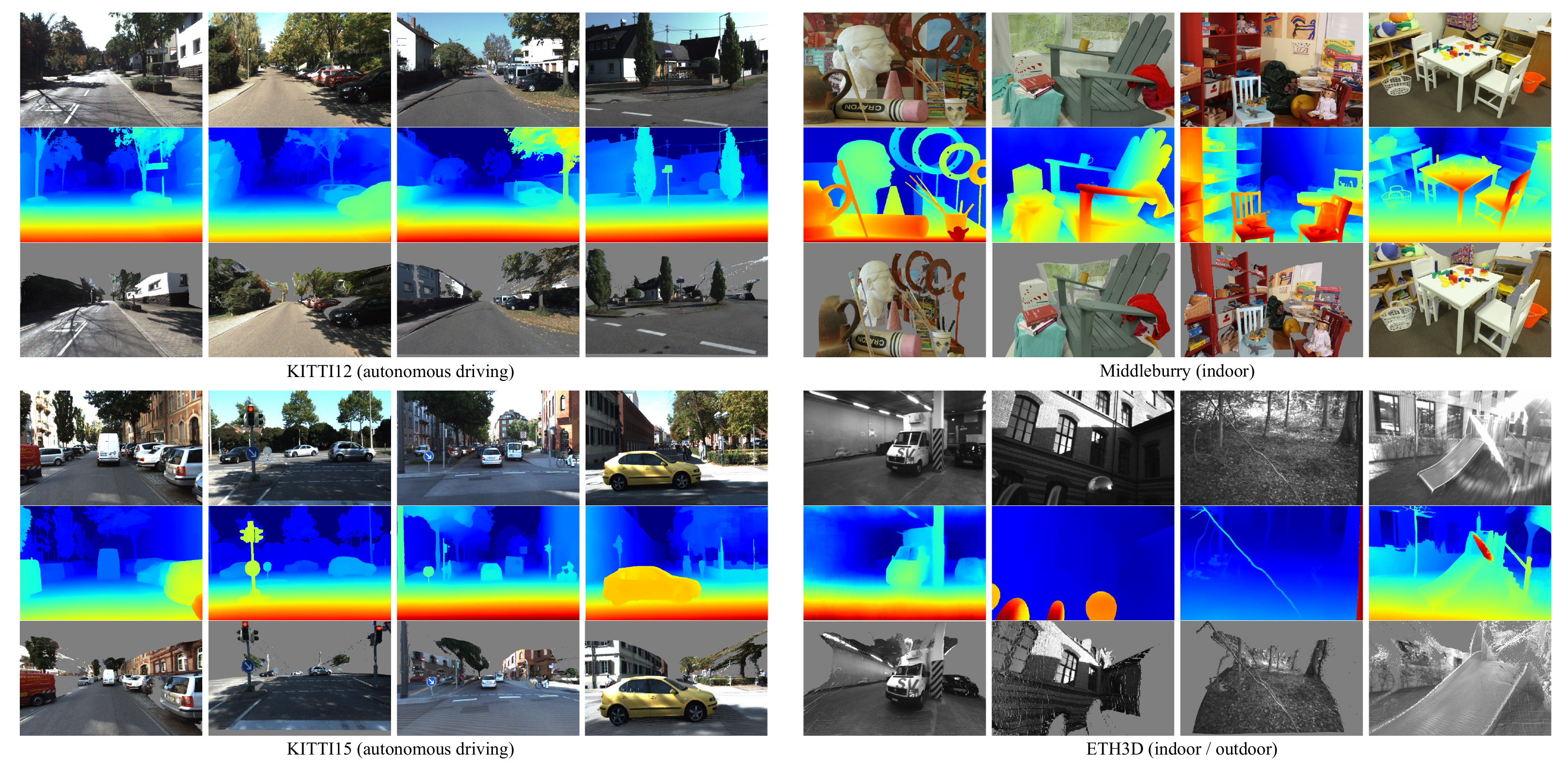}  %
    \caption{StereoAnything demonstrates impressive generalization capabilities in diverse scenarios, including autonomous driving, indoor/outdoor, and medical images. Our final dataset enables training any stereo model to achieve stronger zero-shot generation capabilities.}
    \label{fig:finalvis}  
\end{figure*}

Table~\ref{tab:combine} reports the cross-domain evaluation results when combining mixed labeled stereo datasets with additional pseudo-stereo data. The baseline MIX10 setting already achieves strong performance with a mean error of 3.83, outperforming individual pseudo-stereo data sources such as Obj~\cite{shao2019objects365}. When further incorporating Obj into the MIX10 corpus (MIX10+Obj), we observe consistent improvements on KITTI15 and ETH3D, with errors reduced from 3.56 to 3.53 and from 2.10 to 1.85, respectively. Although the performance on Middlebury slightly decreases, the overall mean remains competitive at 3.96, close to the best baseline. These results highlight that pseudo-stereo datasets can complement mixed labeled stereo data by enriching scene diversity, particularly benefiting challenging benchmarks with diverse domains such as KITTI15 and ETH3D.

\begin{figure}[t]  
    \centering 
    \includegraphics[width=0.48\textwidth]{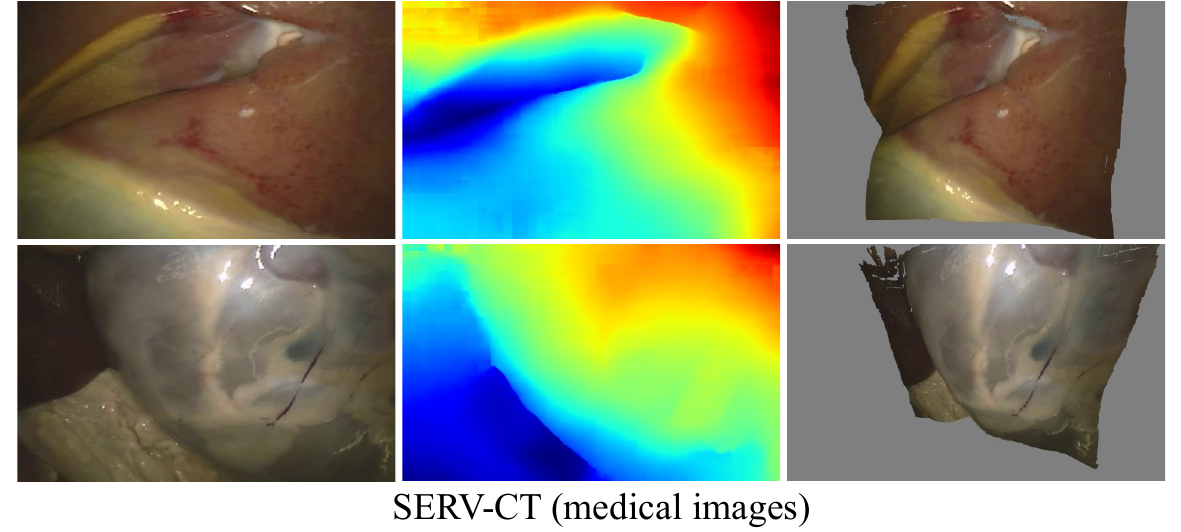}  %
    \caption{\textbf{Visualization on the SERV-CT dataset.} The first column shows the original RGB images, while the second column displays the predicted depth maps. The last column provides the RGB point cloud visualization, where disparity is transformed into a 3D point cloud.}
    \label{fig:serv-ct}  
\end{figure}

\textbf{Qualitative Results.}
In Fig.~\ref{fig:finalvis}, we present the more qualitative results of Stereo Anything on four previously unseen datasets. 
StereoAnything demonstrates strong robustness across various domains, including indoor and outdoor scenes. In addition, Fig.~\ref{fig:serv-ct} shows the visualization results on the SERV-CT dataset, which consists of medical images. These results highlight the effectiveness of our model in accurately estimating depth information from medical imagery, as evidenced by the fine details preserved in the depth maps and the smooth transitions observed in the corresponding point cloud visualizations. The SERV-CT dataset, with its complex medical imagery, further emphasizes the model's robustness in handling intricate, textureless regions and occlusions.

\section{Conclusion}

In this paper, we introduce StereoAnything, a highly practical solution for robust stereo matching, designed to enhance model generalization across a variety of domains and real-world scenarios. We focus on the importance of training with diverse data sources, combining labeled stereo datasets with pseudo-stereo datasets generated via monocular depth estimation models. By leveraging this hybrid approach, we provide a robust training framework capable of handling the inherent challenges of real-world environments.
Our extensive experiments demonstrate that the quality and diversity of training datasets are pivotal to achieving superior performance in stereo matching tasks. Specifically, the inclusion of pseudo-stereo datasets generated from monocular depth maps significantly boosts the model's ability to generalize across unseen environments.
These results underscore the effectiveness of hybrid training strategies, where the combination of curated labeled datasets and large-scale pseudo-stereo data enables models to learn more comprehensive geometric priors, ensuring improved performance and generalization. Our findings also highlight the essential role of data diversity in training deep learning models for stereo matching, particularly when dealing with complex, real-world imagery that traditional datasets alone cannot fully represent.

Looking ahead, there are several promising avenues for future work. First, the expansion of pseudo-stereo data generation techniques, particularly those that incorporate more sophisticated monocular depth estimation models, could further improve the quality and diversity of the training data. Additionally, integrating advanced techniques like domain adaptation and self-supervised learning could further reduce the dependency on labeled data, making the approach more scalable and applicable to even larger, more diverse datasets. Finally, exploring the integration of our method with other computer vision tasks, such as 3D scene reconstruction and multi-view stereo, could open new frontiers for practical applications in autonomous driving, robotics, and AR/VR technologies.

\textbf{Acknowledgements.} This work was supported by the National Natural Science Foundation of China under Grant 62373356 and the Joint Funds of the National Natural Science Foundation of China under U24B20162.


{\small
\bibliographystyle{ieee}
\bibliography{main}
}

\end{document}